\PassOptionsToPackage{table,dvipsnames}{xcolor} %
\documentclass{article} %
\usepackage{iclr2020_conference,times}
\iclrfinalcopy
\usepackage{xcolor}
\definecolor{deepblue}{RGB}{0, 47, 85}

\usepackage[table]{xcolor}
\usepackage{tabularx}
\usepackage{booktabs}
\usepackage{multirow}

\usepackage{amsmath}
\usepackage{amsfonts}
\usepackage{nicefrac} %
\usepackage{cases}
\usepackage{bm}

\usepackage[utf8]{inputenc}
\usepackage[T1]{fontenc}
\usepackage{pifont}
\usepackage{xspace}
\usepackage[none]{hyphenat} %
\usepackage{listings}
\usepackage[normalem]{ulem}
\usepackage[para]{footmisc}

\usepackage[pagebackref=true,colorlinks=true,citecolor=deepblue,linkcolor=deepblue, urlcolor=deepblue]{hyperref}

\usepackage{tikz}
\usepackage{wrapfig}
\usetikzlibrary{external,positioning}
\usepackage{pgfplots}
\pgfplotsset{compat=1.14, legend style={font=\tiny},label style={font=\small}, tick label style={font=\small},}
\usetikzlibrary{matrix}
\usepgfplotslibrary{fillbetween, groupplots}

\usepackage{algorithm}
\usepackage[noend]{algpseudocode}

\definecolor{cycle1}{RGB}{64, 83, 211}
\definecolor{cycle2}{RGB}{221, 179, 16}
\definecolor{cycle3}{RGB}{181, 29, 20}
\definecolor{cycle4}{RGB}{0, 190, 255}
\definecolor{cycle5}{RGB}{251, 73, 176}
\definecolor{cycle6}{RGB}{0, 178, 93}
\definecolor{cyclegray}{RGB}{202, 202, 202}

\newcommand*\samethanks[1][\value{footnote}]{\footnotemark[#1]}

\newcommand{\thiswork}{\textsc{IMD}\xspace}

\newtheorem{theorem}{Theorem}
\newtheorem{definition}{Definition}

\newcommand*{\bigO}{\mathcal{O}}
\newcommand*{\hkt}{\mathrm{hkt}}
\newcommand*{\knn}{k\mathrm{NN}}

\DeclareMathOperator{\ke}{k}
\DeclareMathOperator{\tr}{Tr}

\DeclareMathOperator{\pr}{Pr}

\newcommand*{\vV}{\mathbf{v}}
\newcommand*{\vU}{\mathbf{u}}
\newcommand*{\vQ}{\mathbf{q}}
\newcommand*{\vE}{\mathbf{e}}
\newcommand*{\vZ}{\mathbf{z}}

\newcommand*{\mA}{\mathbf{A}}

\newcommand*{\mD}{\mathbf{D}}
\newcommand*{\mH}{\mathbf{H}}
\newcommand*{\mI}{\mathbf{I}}
\newcommand*{\mJ}{\mathbf{J}}
\newcommand*{\mL}{\bm{\mathcal{L}}}
\newcommand*{\mQ}{\mathbf{Q}}
\newcommand*{\mT}{\mathbf{T}}
\newcommand*{\mU}{\mathbf{U}}
\newcommand*{\mPhi}{\bm{\Phi}}
\newcommand*{\mLambda}{\bm{\Lambda}}
\newcommand*{\mX}{\mathbf{X}}
\newcommand*{\mY}{\mathbf{Y}}

\newcommand*{\pP}{\mathbb{P}}

\newcommand*{\cD}{\mathbb{\mathcal{D}}}
\newcommand*{\cM}{\mathcal{M}}
\newcommand*{\cN}{\mathcal{N}}
\newcommand*{\cX}{\mathcal{X}}

\newcommand*{\sR}{\mathbb{R}}

\newcommand{\mpara}[1]{\medskip\noindent{\bf #1}}
\newcommand{\para}[1]{\noindent{\bf #1}}

\newcommand{\pk}[1]{{#1}} %
\newcommand{\pki}[1]{{#1}} %
\newcommand{\dm}[1]{#1}
\newcommand{\dmn}[1]{{{#1}}}
\newcommand{\ants}[1]{{#1}} %
\newcommand{\mm}[1]{{#1}} %

\pgfplotsset{
  /pgfplots/line legend with two lines/.style 2 args={
    legend image code/.code={
      \draw[mark phase=2, mark repeat=2, #1] plot coordinates {(0cm, -0.1cm) (0.25cm, -0.1cm) (0.5cm, -0.1cm)};
      \draw[mark phase=2, mark repeat=2, #2] plot coordinates {(0cm, 0.1cm) (0.25cm, 0.1cm) (0.5cm, 0.1cm)};
    }
  }
}

\definecolor{codegreen}{rgb}{0,0.6,0}
\definecolor{codegray}{rgb}{0.5,0.5,0.5}
\definecolor{codepurple}{rgb}{0.58,0,0.82}
\definecolor{backcolour}{rgb}{0.95,0.95,0.92}
 
\lstdefinestyle{mystyle}{
    backgroundcolor=\color{white},   
    commentstyle=\color{cycle6},
    keywordstyle=\color{cycle3},
    stringstyle=\color{cycle1},
    basicstyle=\ttfamily\footnotesize,
    breakatwhitespace=false,         
    breaklines=false,                 
    captionpos=b,                    
    keepspaces=true,                 
    numbers=none,                    
    numbersep=5pt,                  
    showspaces=false,                
    showstringspaces=false,
    showtabs=false,                  
    tabsize=2
}
 
\title{The Shape of Data:\\Intrinsic Distance for Data Distributions} %

\author{%
  Anton Tsitsulin\thanks{Equal contribution. Contact~\texttt{tsitsulin@bit.uni-bonn.de},~\texttt{marina.munkhoeva@skoltech.ru}}\;\thanks{University of Bonn}
\And
Marina Munkhoeva\samethanks[1]\;\thanks{Skoltech}
\And
  Davide Mottin\thanks{Aarhus University}
\And
  Panagiotis Karras\samethanks[4]
\And
  Alex Bronstein\thanks{Technion}
\And
  Ivan Oseledets\samethanks[3]
\And
  Emmanuel M{\"u}ller\samethanks[2]
} 
\begin{document}

\maketitle

\begin{abstract}

\pki{The ability to represent and compare machine learning models is crucial in order to quantify subtle model changes, evaluate generative models, and gather insights on neural network architectures.}
Existing techniques for comparing \ants{data distributions} focus on global data properties such as mean and covariance; \pk{in that sense, they are \emph{extrinsic} and \emph{uni-scale}}. We develop a \pki{first-of-its-kind} \emph{intrinsic} and \emph{multi-scale} method for characterizing and comparing data manifolds,
using a lower-bound of the spectral Gromov-Wasserstein inter-manifold distance, \pk{which compares \emph{all} data moments}.
In a thorough experimental study, we demonstrate that
\ants{our method effectively discerns the structure of data manifolds even on unaligned data of different dimensionality,}
\ants{and showcase its efficacy in evaluating the quality of generative models.}

\mm{
}
\end{abstract}

\section{Introduction}\label{sec:introduction}

\dm{The geometric properties of neural networks provide insights about their internals~\citep{morcos2018, wang2018} and help researchers in the design of more robust models~\citep{arjovsky2017,binkowski2018}.}
\ants{Generative models are a natural example of the need for geometric comparison of distributions.
As generative models aim to reproduce the true data distribution $\pP_d$ by means of the model distribution $\pP_g(\vZ;\Theta)$, more delicate evaluation procedures are needed.
}
\ants{Oftentimes, we wish to compare data lying in entirely different spaces, for example to track model evolution or compare models having different representation space.}
\begin{wrapfigure}[16]{r}{5cm}
\vspace{-9mm}
\centering
\begin{tikzpicture}
    \begin{axis}[xtick=\empty,ytick=\empty,width=6cm,height=5.5cm,hide axis]
        \addplot[
            scatter/classes={
                a={mark=asterisk,cycle6,opacity=0.45},
                b={mark=*,cycle3, opacity=0.8}
                },
                scatter, only marks,
                scatter src=explicit symbolic]
         table[x=x,y=y,meta=class]
            {data/inceptionsaurus.dat}; %
    \end{axis}
\end{tikzpicture}
\vspace*{-2.5mm}
\caption{Two distributions having the same first 3 moments, meaning FID and KID scores are close to 0.}\label{fig:fid-kid-counterexample}
\end{wrapfigure}
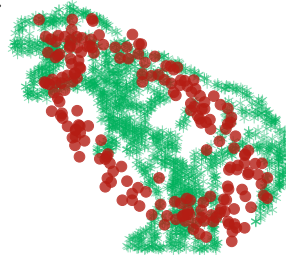 %

In order to evaluate the performance of generative models, past research has proposed several \emph{extrinsic} evaluation measures, most notably the Fr{\'e}chet~\citep{heusel2017} and Kernel~\citep{binkowski2018} Inception Distances (FID and KID).
Such measures only reflect the first two or three moments of distributions, meaning they can be insensitive to global structural problems.
We showcase this inadvertence in Figure~\ref{fig:fid-kid-counterexample}: here FID and KID are insensitive to the global structure of the data distribution.
Besides, as FID and KID are based only on \emph{extrinsic} properties they are unable to compare \emph{unaligned} data manifolds.
\pki{In this paper, we start out from the observation that models capturing the \emph{multi-scale} nature of the data manifold by utilizing higher distribution moment \ants{matching}, such as MMD-GAN~\citep{li2017} and Sphere-GAN~\citep{park2019}, perform consistently better than their single-scale counterparts. On the other hand, using \emph{extrinsic} information can be misleading, as it is dependent on factors external to the data, such as representation. To address this drawback,}
we propose \thiswork, an Intrinsic Multi-scale Distance, 
that is able to compare distributions using \ants{only \emph{intrinsic} information about the data,
and provide an efficient approximation thereof that renders computational complexity nearly linear.}
\ants{We demonstrate that \thiswork effectively quantifies difference in data distributions 
in three distinct application scenarios: comparing word vectors in languages with \emph{unaligned} vocabularies,
tracking dynamics of intermediate neural network representations,
and evaluating generative models.}

\section{Related Work}\label{sec:related-work}

The geometric perspective on data is %
ubiquitous in machine learning. Geometric techniques enhance unsupervised and semi-supervised learning, generative and discriminative models~\citep{belkin2002,arjovsky2017,memoli2011}. We outline the applications of the proposed manifold comparison technique and highlight the geometric intuition along the way.

\subsection{Generative Model Evaluation}\label{ssec:generative-models-eval}

Past research has explored many different directions for the evaluation of generative models. Setting aside models that ignore the true data distribution, such as the Inception Score~\citep{salimans2016} and GILBO~\citep{alemi2018}, we discuss most relevant geometric ideas below; we refer the reader to~\citet{borji2019} for a comprehensive survey.

\mpara{Critic model-based metrics.}
Classifier two-sample tests (C2ST)~\citep{lopezpaz2017} aim to assess whether two samples came from the same distribution by means of an auxiliary classifier.
This idea \pk{is reminiscent of} %
the GAN discriminator network~\citep{goodfellow2014}: if it is possible to train a model \pk{that distinguishes between} samples from the model and the data distributions, \pk{it follows that} these distributions are not entirely similar.
The convergence process of the GAN-like discriminator~\citep{arjovsky2017,binkowski2018}
\pk{lends itself to} %
creating a family of metrics based on training a discriminative classifier~\citep{im2018}. %
Still, training a separate \pk{critic} model is often \pk{computationally} prohibitive %
and requires careful specification. %
Besides, if the critic model is a neural network, \pk{the resulting metric lacks} interpretability and training stability.
Many advanced GAN models such as Wasserstein, MMD, Sobolev and Spherical GANs impose different constraints on the function class so as to stabilize training~\citep{arjovsky2017, binkowski2018, mroueh2017, park2019}. Higher-order moment matching~\citep{binkowski2018,park2019} \pk{enhances GAN performance, enabling GANs to capture multi-scale data properties, while multi-scale noise ameliorates GAN convergence problems~\citep{jenni2019}.} Still, no feasible multi-scale \pk{GAN evaluation} metric has been proposed \pk{to date}.

\mpara{Positional distribution comparison.}
In certain settings, it is acceptable to assign zero probability mass to the real data points~\citep{odena2018}.
\pk{In effect, metrics that} estimate a distribution's location and dispersion provide useful input for generative model evaluations.
For instance, the Fr{\'e}chet Inception Distance (FID)~\citep{heusel2017} computes the Wasserstein-2 (i.e., Fr{\'e}chet) distance between distributions approximated with Gaussians, using only the estimated mean and covariance matrices; the Kernel Inception Distance (KID)~\citep{binkowski2018} computes a polynomial kernel $\ke(x, y) = (\frac{1}{d} x^\top y + 1)^3$ and measures the associated Kernel Maximum Mean Discrepancy (kernel MMD).
Unlike FID, KID has an unbiased estimator~\citep{gretton2012, binkowski2018}. However, even while \pk{such} methods, based on a limited number of moments, \pk{may be} computationally inexpensive, they \pk{only} provide a \pk{rudimentary} characterization of distributions from a geometric viewpoint.

\mpara{Intrinsic geometric measures.}
The Geometry Score~\citep{khrulkov2018} characterizes distributions in terms of their estimated persistent homology, \pk{which roughly corresponds to the number of holes in a manifold}.
Still, the Geometry Score assesses distributions \pk{merely} in terms of their \emph{global} \pk{geometry}. \pk{In this work, we aim to provide a \emph{multi-scale} geometric assessment.}

\subsection{Similarities of Neural Network Representations}\label{ssec:representation-similarity}

\pk{Learning how representations evolve during training or across initializations provides a pathway to the interpretability of neural networks~\citep{raghu2017}. Still, state-of-the-art methods for comparing representations of neural networks~\citep{kornblith2019, morcos2018, wang2018} consider only linear projections. The intrinsic nature of \thiswork renders it appropriate for the task of comparing \pki{neural network} representations, which can only rely on intrinsic information.}

\pki{
\citet{yin2018dimensionality} introduced the Pairwise Inner Product (PIP) loss, an unnormalized covariance error between sets, as a dissimilarity metric between word2vec embedding spaces with common vocabulary.
We show in Section~\ref{ssec:wordvec-dim} how \thiswork is applicable to this comparison task too. %
}
\section{Multi-Scale Intrinsic Distance}

At the core of deep learning lies the \emph{manifold hypothesis}, which states that high-dimensional data, such as images or text, lie on a low-dimensional manifold~\citep{narayanan2010, belkin2002, belkin2007}. 
\pk{We aim to provide a theoretically motivated comparison of data manifolds based on rich intrinsic information.}
\pk{Our target measure should have the following properties:}

\textbf{intrinsic} -- it is invariant to isometric transformations of the manifold, e.g. translations or \pk{rotations}. %

\textbf{multi-scale} -- it captures both local and global information.

\pk{We expose our method starting out with heat kernels, which admit a notion of manifold metric and can be used to lower\hyp{}bound the distance between manifolds.}

\subsection{Heat Kernels on Manifolds and Graphs}\label{ssec:heat-kernel}

\pk{Based on the heat equation, the heat kernel %
captures \emph{all} the information about a manifold's intrinsic geometry~\citep{sun2009}.} %
Given the Laplace-Beltrami operator (LBO) $\Delta_\cX$ on a manifold $\cX$, the \emph{heat equation} is $\frac{\partial u}{\partial t} = \Delta_\cX u$ for $u: \sR^+\!\!\times\!\cX\rightarrow\sR^+$.
A smooth function $u$ is a \emph{fundamental solution} of the heat equation at point $x\in\cX$ if $u$ satisfies both the heat equation and the Dirac condition $u(t, x')\rightarrow\delta(x'\!-x)$ as $t\rightarrow0^+$. \ants{We assume the Dirichlet boundary condition $u(t, x)=0$ for all $t$ and $x\in\partial \cX$.} The heat kernel $\ke_\cX\!\!: \cX\!\times\!\cX\!\times\sR^+\!\!\rightarrow\sR^+_0$ is \pk{the} unique solution of the heat equation; \pk{while heat kernels can be defined} on hyperbolic spaces and other exotic geometries,
we restrict our exposition to Euclidean spaces $\cX=\sR^d$, on which the heat kernel is defined as:
\begin{equation}\label{eq:heat-kernel-rd}
	\ke_{\sR^d}(x, x'\!, t) = \frac{1}{(4\pi t)^{\nicefrac{d}{2}}}\exp\left(-\frac{\lVert x - x' \rVert^2}{4t}\right)%
\end{equation}
For a compact $\cX$ including submanifolds of $\sR^d$, the heat kernel admits the expansion $\ke_\cX(x, x'\!, t) = \sum_{i=0}^\infty e^{-\lambda_it}\phi_i(x)\phi_i(x')$, where $\lambda_i$ and $\phi_i$ are the $i$-th eigenvalue and eigenvector of $\Delta_\cX$. \pk{For $t \simeq 0^+$, according to Varadhan's lemma, the heat kernel approximates geodesic distances.}
\pki{Importantly for our purposes, the Heat kernel is \emph{multi-scale}: 
for a local domain $\cD$ with Dirichlet condition, the localized heat kernel $\ke_\cD(x, x'\!, t)$ is a good approximation of $\ke_\cX(x, x'\!, t)$ if either (i) $\cD$ is arbitrarily small and $t$ is small enough, or (ii) $t$ is for arbitrarily large and $\cD$ is big enough. Formally,
}

\begin{definition}
{\normalfont\textbf{Multi-scale property}}
\citep{grigoryan2006,sun2009}
\ants{(i) For any smooth and relatively compact domain $\cD\subseteq\cX, \lim_{t\rightarrow0}\ke_\cD(x, x'\!, t)=\ke_\cX(x, x'\!, t)$ (ii) For any $t \in \sR^+$ and any $x,x'\!\in \cD_1$ localized heat kernel $\ke_{\cD_1}(x, x'\!, t)\leq\ke_{ \cD_2}(x, x'\!, t)$ if $\cD_1\subseteq\cD_2$. Moreover, if $\{\cD_n\}$ is an expanding and exhausting sequence $\bigcup_{i=1}^{\infty}\cD_i=\cX$ and $\cD_{i-1}\subseteq\cD_{i}$, then $\lim_{i\rightarrow\infty}\ke_{\cD_i}(x,x'\!,t)=\ke_\cX(x, x'\!, t)$ for any $t$.}
\end{definition}

Heat kernels are also defined for graphs in terms of their Laplacian matrices. An undirected graph is a pair $G=(V, E)$, where $V = (v_1, \ldots, v_n), n = |V|$, is the set of vertices and $E\subseteq (V \times V)$ the set of edges.
The \emph{adjacency matrix} of $G$ is a $n\times n$ matrix $\mA$ having $\mA_{ij} \!=\! 1$ if $(i,j) \in E$ and $A_{ij}\! =\! 0$ otherwise.
The \emph{normalized graph Laplacian} is the matrix $\mL\! =\! \mI\! -\! \mD^{-\frac{1}{2}}\mA\mD^{-\frac{1}{2}}$, where $\mD$ is the diagonal matrix in which entry $\mD_{ii}$ holds the degree of node $i$, i.e, $\mD_{ii} = \sum_{j = 1}^n \mA_{ij}$.
Since the Laplacian matrix is symmetric, its eigenvectors $\phi_1, \ldots, \phi_n$, are real and orthonormal.
Thus, it is factorized as $\mL = \mPhi\mLambda\mPhi^\top$, where $\mLambda$ is a diagonal matrix with the sorted eigenvalues $\lambda_1 \le \ldots \le \lambda_n$, and $\Phi$ is the orthonormal matrix $\mPhi = (\phi_1,  \dots, \phi_n)$ having the eigenvectors of $\mL$ as its columns.
\pk{The heat kernel on a graph is also given by the} solution to the heat equation on a graph, which requires an eigendecomposition of its Laplacian: $\mH_t=e^{-t\mL}=\mPhi e^{-t\mLambda}\mPhi^\top = \sum_i e^{-t\lambda_i}\phi_i\phi_i^\top$.

A useful invariant of the heat kernel is the \emph{heat kernel trace} $\hkt_\cX\!: \cX\!\times\!\sR^+_0\!\!\rightarrow \sR^+_0$, defined \pk{by a diagonal restriction as} $\hkt_\cX(t)=\int_{\cX}\ke_\cX(x,x,t)dx=\sum_{i=0}^\infty{e^{-\lambda_it}}$ or, in \pk{the} discrete case, $\hkt_{\mL}(t)=\tr(\mH_t)=\sum_ie^{-t\lambda_i}$. Heat kernels traces (HKTs) have been successfully applied to the analysis of 3D shapes~\citep{sun2009} and graphs~\citep{tsitsulin2018}. The HKT contains \emph{all} the information in the graph's spectrum, both local and global, as the eigenvalues $\lambda_i$ can be \pk{inferred therefrom}~\citep[Remark~4.8]{memoli2011}.
For example, if there are $c$ connected components in the graph, then $\lim_{t\rightarrow\infty}\hkt_{\mL}(t) = c$.

\subsection{Convergence \pk{to the} Laplace-Beltrami Operator}\label{ssec:lbo-convergence}

An important property of graph Laplacians is that it is possible to construct a graph \pk{among points sampled from a manifold $\cX$} such that \pk{the spectral properties of its Laplacian resemble} those of the Laplace-Beltrami operator on $\cX$.
Belkin and Niyogi~\citep{belkin2002} proposed \pk{such a construction,} the point cloud Laplacian, which is used for dimensionality reduction in a technique called Laplacian eigenmaps. Convergence to the LBO has been proven for various definitions of the graph Laplacian, including the one \pk{we use}~\citep{belkin2007, hein2007, coifman2006, ting2010}. We recite the convergence results for the point cloud Laplacian from~\citet{belkin2007}: %
\begin{theorem}\label{thm:belkin-eigenconvergence}
	Let $\lambda_{n,i}^{t_n}$ and $\phi_{n,i}^{t_n}$ be the $i^\mathsf{th}$ eigenvalue and eigenvector, respectively, of the point cloud Laplacian $\mL^{t_n}$; let $\lambda_{i}$ and $\phi_{i}$ be the $i^\mathsf{th}$ eigenvalue and eigenvector of the LBO $\Delta$. Then, there exists $t_n \rightarrow 0$ such that
\begin{equation*}
\begin{split}
	&\lim_{n\rightarrow\infty}\lambda_{n,i}^{t_n} = \lambda_i \\
	&\lim_{n\rightarrow\infty}\left\lVert\phi_{n,i}^{t_n} - \phi_i\right\rVert_2 = 0
\end{split}
\end{equation*}
\end{theorem}
\pk{Still, the point cloud Laplacian involves the creation of an $\bigO(n^2)$ matrix; for the sake of scalability, we use the $k$-nearest-neighbours ($\knn$) graph by OR-construction (i.e., based on bidirectional $\knn$ relationships among points),
whose Laplacian converges to the LBO for data with sufficiently high intrinsic dimension~\citep{ting2010}}. %
As for the choice of $k$, a random geometric $\knn$ graph is connected when $k \geq \nicefrac{\log n}{\log{7}} \approx 0.5139 \log n$~\citep{balister2005};
$k=5$ yields connected graphs for all sample sizes we tested.

\subsection{Spectral Gromov-Wasserstein Distance}\label{ssec:gromov-wasserstein}

\pk{Even while it is} a multi-scale metric \emph{on} manifolds, the heat kernel can be spectrally approximated by finite graphs constructed from points sampled from these manifolds. In order to construct a metric \emph{between} manifolds, \citet{memoli2011} suggests an optimal\hyp{}transport\hyp{}theory\hyp{}based ``meta-distance'': a spectral definition of the Gromov\hyp{}Wasserstein distance between Riemannian manifolds based on matching the heat kernels at all scales.
The cost of matching a pair of points $(x,x')$ on manifold $\cM$ to a pair of points $(y,y')$ on manifold $\cN$ at scale $t$ is given by their heat kernels $\ke_\cM, \ke_\cN$:
\begin{equation*}
\Gamma(x, y, x'\!, y'\!, t) = \left| \ke_\cM(x, x'\!, t) - \ke_\cN(y,y'\!, t) \right|.
\end{equation*}
The distance between the manifolds is then defined in terms of the infimal measure coupling 
\begin{equation*}
d_{\mathrm{GW}}(\cM,\cN) = \inf_{\mu} \sup_{t>0} e^{-2(t+t^{-1})} \, \| \Gamma \|_{L^2(\mu \times \mu)},
\end{equation*}
where the infimum is sought over all measures $\mu$ on $\cM \times \cN$ marginalizing to the standard measures on $\cM$ and $\cN$. For finite spaces, $\mu$ is a doubly-stochastic matrix. This distance is lower\hyp{}bounded~\citep{memoli2011} in terms of the respective heat kernel traces as:
\begin{equation}\label{eq:gw-lowerbound}
d_{\mathrm{GW}}(\cM,\cN)  \ge \sup_{t>0} e^{-2(t+t^{-1})} \, \left| \hkt_\cM(t) - \hkt_\cN(t) \right|.
\end{equation}
This lower bound is the scaled $L_\infty$ distance between the \emph{heat trace signatures} $\hkt_\cM$ and $\hkt_\cN$. The scaling factor $e^{-2(t+t^{-1})}$ favors medium-scale differences, meaning that this lower bound is not sensitive to local perturbations. The maximum of the scaling factor occurs at $t=1$, and \pk{more than $1-10^{-8}$} of the function mass lies between $t=0.1$ and $t=10$.

\subsection{Heat Trace Estimation}\label{ssec:computation}

Calculating the heat trace signature efficiently and accurately is a challenge on a large graph as it involves computing a trace of a large matrix exponential, i.e. $\tr(e^{-t\mL})$. 
A naive approach would be to use an eigendecomposition $\exp(-t\mL) = \Phi\exp(-t\Lambda)\Phi^\top$, which is infeasible for large $n$.
Recent work~\citep{tsitsulin2018} suggested using either truncated Taylor expansion or linear interpolation of the interloping eigenvalues, however, both techniques are quite coarse. 
To combine accuracy and speed, we use the Stochastic Lanczos Quadrature (SLQ)~\citep{ubaru2017, golub2009}.
This method combines the Hutchinson trace estimator~\citep{hutchinson1989, adams2018} and the Lanczos algorithm for eigenvalues.
We aim to estimate the trace of a matrix function with a Hutchinson estimator: \begin{equation}\label{eq:trace}
 	\tr(f(\mL)) = \mathbb{E}_{p(\vV)}(\vV^\top  f(\mL) \vV ) \approx \frac{n}{n_v} \sum_{i=1}^{n_v} \vV_i^\top f(\mL) \vV_i,
 \end{equation}
where the function of interest $f(\cdot) = \exp(\cdot)$ and $\vV_i$ are $n_v$ random vectors drawn from a distribution $p(\vV)$ with zero mean and unit variance.
A typical choice for $p(\vV)$ is Rademacher or a standard normal distribution. In practice, there is little difference, although in theory Rademacher has less variance, but Gaussian requires less random vectors \citep{avron2011}.

To estimate the quadratic form $\vV_i^\top f(\mL) \vV_i$  in \eqref{eq:trace} with a symmetric real-valued matrix $\mL$ and a smooth function $f$,
we plug in the eigendecomposition $\mL = \Phi\Lambda\Phi^\top$,
rewrite the outcome as a Riemann-Stieltjes integral and \pki{apply} the $m$-point Gauss quadrature rule~\citep{golub1969}:
\begin{equation}
\vV_i^\top f(\mL) \vV_i = \vV_i^\top  \Phi f(\Lambda)\Phi^\top \vV_i = \sum\limits_{j=1}^n f(\lambda_j)\mu_j^2 = \int_a^b f(t) d\mu(t) \approx \sum\limits_{k=1}^m \omega_k f(\theta_k),
\end{equation}
where
$\mu_j = [\Phi^\top \vV_i]_j$ and $\mu(t)$ is a piecewise constant function defined as follows
$$\mu(t) = \begin{cases}
			0, &\text{if } t<a = \lambda_n \\
			\sum_{j=1}^i \mu_j^2, &\text{if } \lambda_i\leq t < \lambda_{i-1} \\
			\sum_{j=1}^n \mu_j^2, &\text{if } b=\lambda_1\leq t
		   \end{cases}
$$
and $\theta_k$ are the quadrature's nodes and $\omega_k$ are the corresponding weights.
We obtain $\omega_k$ and $\theta_k$ with the $m$-step Lanczos algorithm~\citep{golub2009},
\pki{which we describe} succinctly.

Given the symmetric matrix $\mL$ and an arbitrary \emph{starting unit-vector} $\vQ_0$, the $m$-step Lanczos algorithm computes an $n \times m$ matrix $\mQ = [\vQ_0, \vQ_1, \dots, \vQ_{m-1}]$ with orthogonal columns and an $m \times m$ tridiagonal symmetric matrix $\mT$, such that $\mQ^\top \mL \mQ = \mT$.
The columns of $\mQ$ constitute an orthonormal basis for the Krylov subspace $\mathcal{K}$ that spans vectors $\{\vQ_0, \mL \vQ_0, \dots, \mL^{m-1}\vQ_0\}$; each $\vQ_i$ vector is given as a polynomial in $\mL$ applied to the initial vector $\vQ_0$: $\vQ_i = p_i(\mL)\vQ_0$.
These Lanczos polynomials are orthogonal with respect to the integral measure $\mu(t)$.
As orthogonal polynomials satisfy the three term recurrence relation, we obtain $p_{k+1}$ as a combination of $p_k$ and $p_{k-1}$. The tridiagonal matrix storing the coefficients of such combinations, called the Jacobi matrix $\mJ$, is exactly the tridiagonal symmetric matrix $\mT$.
A classic result tells us that the nodes $\theta_k$ and the weights $\omega_k$ of the Gauss quadrature
are the eigenvalues of $\mT$, $\lambda_k$, and the squared first components of its normalized eigenvectors, $\tau_k^2$, respectively (see \citet{golub1969, wilf1962mathematics, golub2009}).
Thereby, setting $\vQ_0 = \vV_i$, the estimate for the quadratic form becomes:
\begin{equation}\label{eq:quadratic}
	\vV_i^\top f(\mL) \vV_i \approx \sum\limits_{k=1}^m \tau_k^2 f(\lambda_k), \quad
	\tau_k = \mU_{0,k} = \vE_1^\top \vU_k, \quad \lambda_k = \Lambda_{k,k} \quad \mT = \mU \Lambda \mU^\top,
\end{equation}
Applying~\eqref{eq:quadratic} over $n_v$ random vectors in the Hutchinson trace estimator~\eqref{eq:trace} yields the SLQ estimate:
\begin{equation}\label{eq:slq}
	\tr(f(\mL)) \approx \frac{n}{n_v} \sum_{i=1}^{n_v} \left(\sum\limits_{k=0}^m \left(\tau_k^i\right)^2 f\!\left(\lambda_k^i\right)\right) = \Gamma.
\end{equation}
We derive error bounds for the estimator based on the Lanczos approximation of the matrix exponential, and show that even a few Lanczos steps, i.e., $m=10$, are sufficient for an accurate approximation of the quadratic form. However, the trace estimation error is theoretically dominated by the error of the Hutchinson estimator, e.g. for Gaussian $p(\vV)$ the bound on the number of samples to guarantee that the probability of the relative error exceeding $\epsilon$ is at most $\delta$ is $8\epsilon^{-2}\ln(2/\delta)$ \citep{roosta2015}. Although, in practice, we observe performance much better than the bound suggests. Hutchinson error implies nearing accuracy roughly $10^{-2}$ with $n_v \geq 10$k random vectors, however, with as much as $n_v=100$ the error is already  $10^{-3}$. Thus, we use default values of $m=10$ and $n_v=100$ in all experiments in Section \ref{sec:experiments}. Please see Appendix A for full derivations and figures.

\subsection{Putting \thiswork{} Together}\label{ssec:thiswork}

We employ the heretofore described advances in differential geometry and numerical linear algebra to create \thiswork{} (\emph{Multi\hyp{}Scale Intrinsic Distance}), a fast, intrinsic method to lower-bound the spectral Gromov-Wasserstein distance between manifolds.

\begin{wrapfigure}[10]{r}{6.5cm}
    \begin{minipage}{6.5cm}
\small
\vspace{-11mm}
     \begin{algorithm}[H]
    \begin{algorithmic}%
    \Function{IMDesc}{$X$} %
        \State{$G \gets \mathtt{kNN}(X)$}
        \State{$\mL \gets \mathtt{Laplacian}(G)$}
        \State{\Return{$\Gamma=\mathtt{slq}(\mL, s, n_v)$}}
    \EndFunction{}
    
    \vspace{0.5em}
    \Function{IMDist}{$X, Y$}
        \State{$\texttt{hkt}_{X} \gets \mathtt{IMDist}(X)$}
        \State{$\texttt{hkt}_{Y} \gets \mathtt{IMDist}(Y)$}
        \State{\Return{$\sup e^{-2(t+t^{-1})} | \texttt{hkt}_{X} - \texttt{hkt}_{Y} |$}}
    \EndFunction{}

    \end{algorithmic}
    \caption{\thiswork algorithm.}\label{alg:ourwork}
\end{algorithm}
\end{minipage}
\end{wrapfigure}

We describe the overall computation of \thiswork{} in Algorithm~\ref{alg:ourwork}. 
Given data samples in $\sR^{d}$, we build a $\knn$ graph $G$ by OR-construction such that its Laplacian spectrum approximates the one of the Laplace-Beltrami operator of the underlying manifold~\citep{ting2010}, and then compute $\hkt_G(t)=\sum_ie^{-\lambda_it}\approx \Gamma$.
We compare heat traces in the spirit of Equation~\eqref{eq:gw-lowerbound}, i.e., $\left| \hkt_{G_1}(t) - \hkt_{G_2}(t) \right|$ for $t\in(0.1, 10)$ sampled from a logarithmically spaced grid.

Constructing exact $\knn$ graphs is an  $\bigO(dn^2)$ operation; however, approximation algorithms take near-linear time $\bigO(dn^{1+\omega})$~\citep{dong2011,aumuller2019}. In practice, with approximate $\knn$ graph construction~\citep{dong2011}, computational time is low while result variance is similar to the exact case.
The $m$-step Lanczos algorithm on a sparse $n \times n$ $\knn$ Laplacian $\mL$ with one starting vector has $\mathcal{O}(knm)$ complexity, where $kn$ is the number of nonzero elements in $\mL$. The symmetric tridiagonal matrix eigendecomposition incurs an additional $\mathcal{O}(m \log m)$~\citep{coakley2013}. We apply this algorithm over $n_v$ starting vectors, yielding a complexity of $\bigO(n_v(m\log m + kmn))$, with constant $k=5$ and $m=10$ by default. In effect, \thiswork{}'s time complexity stands between those of two common GAN evaluation methods: KID, which is $\bigO(dn^2)$ and FID, which is $\bigO(d^3+dn)$. The time complexity of Geometry Score is unspecified in~\citet{khrulkov2018}, yet in Section~\ref{ssec:stability-scalability} we show that its runtime grows exponentially in sample size.

\section{Experiments}\label{sec:experiments}
\setcounter{footnote}{0}
\vspace{-2mm}

\dmn{We evaluate \thiswork{} on the ability to compare intermediate representations of machine learning models. For instance, in a recommender system we could detect whether a problem is related to the representation or the the classifier in the end of a pipeline.  
In this section, we show the effectiveness of our intrinsic measure on multiple tasks and show how our intrinsic distance can provide insights beyond previously proposed extrinsic measures. 
}
\begin{figure*}[t]
	\begin{tikzpicture}
    \begin{groupplot}[group style={
                      group name=myplot,
                      group size= 2 by 1, horizontal sep=3.2cm},height=5cm]
		\nextgroupplot[
			width=5cm,
xticklabels={pl,ru,el,hu,tr,ar,he,en,simple,sv,de,es,nl,pt,vi,war},
yticklabels={pl,ru,el,hu,tr,ar,he,en,simple,sv,de,es,nl,pt,vi,war},
    yticklabel style = {font=\tiny, anchor=east, inner sep=1pt},
    xticklabel style = {rotate=45,font=\tiny, anchor=north east, inner sep=1pt},
	xtick={0,...,15},
    ytick={15,...,0},
    ytick style={draw=none},
    xtick style={draw=none},
    xmin=0,
    xmax=15,
    ymin=0,
    ymax=15,
    enlargelimits={abs=0.5},
    point meta=explicit,
    axis on top,   
    colormap/viridis,
    colorbar,
    point meta min={0},
    point meta max={150},
    colorbar style={
            ytick={0, 50, 100, 150},
            width=0.05*\pgfkeysvalueof{/pgfplots/parent axis width},
        },
 	title style={at={(0.5,0)},anchor=north,yshift=-6mm},
 	title = {a},
		]
		\addplot [matrix plot*,mesh/cols=16,mesh/rows=16] file [meta=index 2] {data/exp/word2vec.langmap.msid.dat};
        \nextgroupplot[
			only marks,
    		legend style={at={(0.2,0.95)},
      anchor=north,legend columns=1},
    		ylabel={Change wrt.\ Simple},
    		xtick=data,
    		xticklabel style = {rotate=45,font=\tiny, anchor=north east, inner sep=1pt},
    		xticklabels from table={data/exp/word2vec.difflangs.english.dat}{lang},
    		width=7.5cm,
    		extra y ticks ={1},
    		extra tick style={grid=major,major grid style={black,thick}},
 	title style={at={(0.5,0)},anchor=north,yshift=-6mm},
 	title = {b},
		]
		\addplot[mark=*,mark options={color=cycle1}, mark size=3pt] table [x expr=\coordindex, y=lsd] {data/exp/word2vec.difflangs.english.dat};\addlegendentry{\thiswork};
\addplot[mark=square*,mark options={color=cycle2}, mark size=3pt] table [x expr=\coordindex, y=fid] {data/exp/word2vec.difflangs.english.dat};\addlegendentry{FID};
\addplot[mark=triangle*,mark options={color=cycle3}, mark size=3pt] table [x expr=\coordindex, y=kid] {data/exp/word2vec.difflangs.english.dat};\addlegendentry{KID};
\draw [<-, very thick, color=black] (1, 2)-- +(0pt,+16pt) node[above,inner sep=0pt,align=left] (english) {\tiny English};
\node[inner sep=0pt, above=1pt of english] (simple) {\tiny Simple};
  \end{groupplot}
\end{tikzpicture}
\vspace*{-5mm}
\caption{(a) \thiswork distances between language pairs for unaligned Wikipedia word embeddings  and (b) distances from the simple English Wikipedia visualized for \thiswork, FID, and KID. We consider 16 languages: Polish, Russian, Greek, Hungarian, Turkish, Arabic, Hebrew, English, Simple English, Swedish, German, Spanish, Dutch, Portugese, Vietnamese, and Waray-Waray.}\label{fig:wordvec-multifig}
\vspace{-4mm}
\end{figure*}
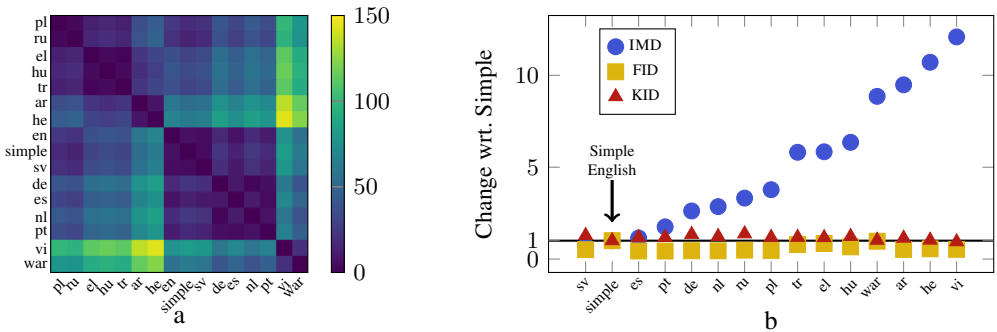 
\dmn{\para{Summary of experiments.} We examine the ability of \thiswork{}\footnote{Our code is available open-source: \url{https://github.com/xgfs/imd}.} to measure several aspects of difference among data manifolds. 
We first consider a task from unsupervised machine translation with unaligned word embeddings and show that \thiswork{} captures correlations among language kinship (affinity or genealogical relationships). Second, we showcase how \thiswork{} handles data coming from data sources of unequal dimensionalities. Third, we study how \thiswork highlights differences among image data representations across initializations and through training process of neural networks.}

\subsection{Comparing Unaligned Language Manifolds}
The \dmn{problem of unaligned representations is particularly severe in the domain of natural language processing as the vocabulary is rarely comparable across different languages or even different documents.}
\dmn{
We employ \thiswork{} to measure the relative closeness of pairs of languages based on the word embeddings with different vocabularies. Figure~\ref{fig:wordvec-multifig}~(a) shows  a heatmap of pairwise \thiswork{} scores. \thiswork{} detects similar languages (Slavic, Semitic, Romanic, etc.) despite the lack of ground truth vocabulary alignment. %
On the other hand, Figure~\ref{fig:wordvec-map-fidkid} in Appendix~\ref{ssec:extra-experiments} shows that FID and KID, are not able to distinguish the intrinsic language-specific structure in word embeddings.
Detailed description and setting of the experiment can be found in Appendix~\ref{ssec:extra-experiments}.
}

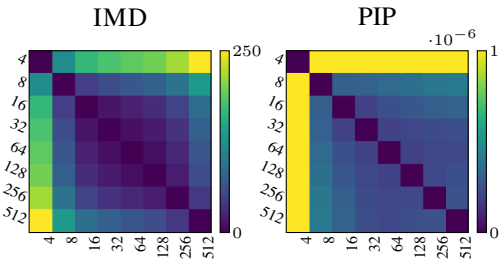
\begin{wrapfigure}[13]{r}{7cm}
\vspace{-8mm}
	\begin{tikzpicture}
    \begin{groupplot}[group style={
                      group name=myplot,
                      group size= 2 by 1, horizontal sep=1cm},width=4cm,
                      height=4cm,
xticklabels={4,8,16,32,64,128,256,512},
yticklabels={4,8,16,32,64,128,256,512},
	yticklabel style = {font=\tiny, inner sep=1pt},
    xticklabel style = {rotate=45,font=\tiny, anchor=north east, inner sep=1pt},
	xtick={0,...,7},
    ytick={7,...,0},
    ytick style={draw=none},
    xtick style={draw=none},
    xmin=0,
    xmax=7,
    ymin=0,
    ymax=7,
    enlargelimits={abs=0.5},
    point meta=explicit,
    axis on top,   
    colormap/viridis,]
\nextgroupplot[
    yticklabel style = {rotate=-25,font=\tiny, anchor=east, inner sep=1pt},
    xticklabel style = {rotate=45,font=\tiny, anchor=north east, inner sep=1pt},
    colormap/viridis,
    colorbar,
    point meta min={0},
    point meta max={250},
    colorbar style={
            ytick={0, 250},
            at={(1.05,1)},
            width=0.05*\pgfkeysvalueof{/pgfplots/parent axis width},
        },
 	title style={at={(0.5,1.2)},anchor=north,yshift=0mm},
 	title = \thiswork,
    ]
\addplot [matrix plot*,mesh/cols=8,mesh/rows=8] file [meta=index 2] {data/exp/word2vec.diffdim.en.dat};		\nextgroupplot[
    yticklabel style = {rotate=-25,font=\tiny, anchor=east, inner sep=1pt},
    xticklabel style = {rotate=45,font=\tiny, anchor=north east, inner sep=1pt},
    colormap/viridis,
    colorbar,
    point meta min={0},
    point meta max={1e-6},
    colorbar style={
            ytick={0, 1e-6},
            at={(1.05,1)},
            width=0.05*\pgfkeysvalueof{/pgfplots/parent axis width},
        },
 	title style={at={(0.5,1.2)},anchor=north,yshift=0mm},
 	title = PIP,
    ]
\addplot [matrix plot*,mesh/cols=8,mesh/rows=8] file [meta=index 2] {data/exp/word2vec.diffdim.covar.en.dat};
  \end{groupplot}
\end{tikzpicture}
\caption{\dmn{Comparison of \thiswork{} and PIP loss on word embeddings of different dimension. \thiswork detects subtle changes in the dimensionality.}}\label{fig:wordvec-map-diffdim-msid-pip}
\end{wrapfigure} \subsection{Optimizing Dimensionality of Word Embeddings}\label{ssec:wordvec-dim}
Comparing \dmn{data having different dimensionality is cumbersome, even when representations \emph{are} aligned.
We juxtapose \thiswork by PIP loss~\citep{yin2018dimensionality} which allows the comparison of aligned representations for word embeddings.}
\dmn{
To this end, we measure \thiswork{} distance between English word embeddings of varying dimensions. Figure~\ref{fig:wordvec-map-diffdim-msid-pip} shows the heatmap of the scores between sets of word vectors of different dimensionalities.
Closer dimensionalities have lower distance scores for both metrics. 
However, \thiswork{} better highlights gradual change of the size of word vectors, e.g.\ word vectors of size 4 and 8 are clearly closer to each other than embeddings of size 4 and 16 in terms of \thiswork{}, which is not true for PIP.
}

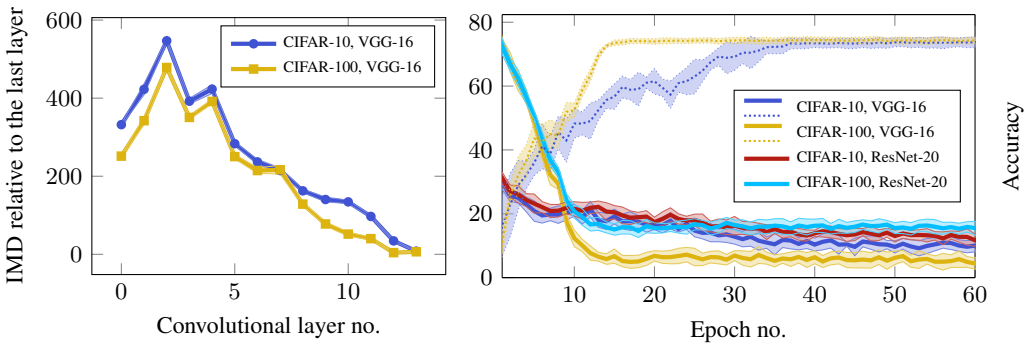
\begin{figure*}[b]
	\begin{tikzpicture}
    \begin{groupplot}[group style={
                      group name=myplot,
                      group size= 2 by 1, horizontal sep=0.75cm},height=5cm, width=0.45\linewidth]
        \nextgroupplot[
			ylabel={\thiswork{} relative to the last layer},
			xlabel={Convolutional layer no.},
			legend cell align=left,
			legend pos = north east,
			]
		\addplot[ultra thick,color=cycle1,mark=*,mark size=1pt] table[x=samples,y=lsd] {data/exp/manifold-stability/manifold-evolution-cifar10-vgg16.dat};
		\addlegendentry{CIFAR-10, VGG-16};
		\addplot[name path=cifar10_top,color=cycle1!70,forget plot] table[x=samples,y=cih] {data/exp/manifold-stability/manifold-evolution-cifar10-vgg16.dat};
		\addplot[name path=cifar10_btm,color=cycle1!70,forget plot] table[x=samples,y=cil] {data/exp/manifold-stability/manifold-evolution-cifar10-vgg16.dat};
		\addplot[cycle1!50,fill opacity=0.5,forget plot] fill between[of=cifar10_top and cifar10_btm];
		\addplot[ultra thick,color=cycle2,mark=square*,mark size=1pt] table[x=samples,y=lsd] {data/exp/manifold-stability/manifold-evolution-cifar100-vgg16.dat};
		\addlegendentry{CIFAR-100, VGG-16};
		\addplot[name path=cifar100_top,color=cycle2!70,forget plot] table[x=samples,y=cih] {data/exp/manifold-stability/manifold-evolution-cifar100-vgg16.dat};
		\addplot[name path=cifar100_btm,color=cycle2!70,forget plot] table[x=samples,y=cil] {data/exp/manifold-stability/manifold-evolution-cifar100-vgg16.dat};
		\addplot[cycle2!50,fill opacity=0.5,forget plot] fill between[of=cifar100_top and cifar100_btm];
		\nextgroupplot[
			scale only axis,
			xlabel={Epoch no.},
			xmin=1,
			xmax=60,
			ymin=0,
			legend cell align=left,
			legend style={at={(0.97,0.5)},anchor=east},
			axis y line*=left,
 			legend entries={{CIFAR-10, VGG-16}, {CIFAR-100, VGG-16}, {CIFAR-10, ResNet-20}, {CIFAR-100, ResNet-20}},
 			height=3.5cm,
		]
		\addlegendimage{line legend with two lines={thick, densely dotted, draw=cycle1}{ultra thick,color=cycle1}};
		\addlegendimage{line legend with two lines={thick, densely dotted, draw=cycle2}{ultra thick,color=cycle2}};
		\addlegendimage{ultra thick,color=cycle3};
		\addlegendimage{ultra thick,color=cycle4};
	
		\addplot[ultra thick,color=cycle1] table[x=samples,y=lsd] {data/exp/manifold-stability/training-evolution-cifar10-vgg16.dat};
		\addplot[name path=cifar10_top,color=cycle1!70,forget plot] table[x=samples,y=lsd_cih] {data/exp/manifold-stability/training-evolution-cifar10-vgg16.dat};
		\addplot[name path=cifar10_btm,color=cycle1!70,forget plot] table[x=samples,y=lsd_cil] {data/exp/manifold-stability/training-evolution-cifar10-vgg16.dat};
		\addplot[cycle1!50,fill opacity=0.5,forget plot] fill between[of=cifar10_top and cifar10_btm];
	
		\addplot[ultra thick,color=cycle2] table[x=samples,y=lsd] {data/exp/manifold-stability/training-evolution-cifar100-vgg16.dat};
		\addplot[name path=cifar100_top,color=cycle2!70,forget plot] table[x=samples,y=lsd_cih] {data/exp/manifold-stability/training-evolution-cifar100-vgg16.dat};
		\addplot[name path=cifar100_btm,color=cycle2!70,forget plot] table[x=samples,y=lsd_cil] {data/exp/manifold-stability/training-evolution-cifar100-vgg16.dat};
		\addplot[cycle2!50,fill opacity=0.5,forget plot] fill between[of=cifar100_top and cifar100_btm];
	
		\addplot[ultra thick,color=cycle3] table[x=samples,y=lsd] {data/exp/manifold-stability/training-evolution-cifar10-resnet20.dat};
		\addplot[name path=cifar10_top_,color=cycle3!70,forget plot] table[x=samples,y=lsd_cih] {data/exp/manifold-stability/training-evolution-cifar10-resnet20.dat};
		\addplot[name path=cifar10_btm_,color=cycle3!70,forget plot] table[x=samples,y=lsd_cil] {data/exp/manifold-stability/training-evolution-cifar10-resnet20.dat};
		\addplot[cycle3!50,fill opacity=0.5,forget plot] fill between[of=cifar10_top_ and cifar10_btm_];
	
		\addplot[ultra thick,color=cycle4] table[x=samples,y=lsd] {data/exp/manifold-stability/training-evolution-cifar100-resnet20.dat};
		\addplot[name path=cifar100_top_,color=cycle4!70,forget plot] table[x=samples,y=lsd_cih] {data/exp/manifold-stability/training-evolution-cifar100-resnet20.dat};
		\addplot[name path=cifar100_btm_,color=cycle4!70,forget plot] table[x=samples,y=lsd_cil] {data/exp/manifold-stability/training-evolution-cifar100-resnet20.dat};
		\addplot[cycle4!50,fill opacity=0.5,forget plot] fill between[of=cifar100_top_ and cifar100_btm_];
		\end{groupplot}
  \begin{groupplot}[group style={
                      group name=myplot,
                      group size= 2 by 1, horizontal sep=0.75cm},height=5cm, width=0.45\linewidth]
        \nextgroupplot[group/empty plot];
        \nextgroupplot[
			scale only axis,
			ylabel={Accuracy},
			xmin=1,
			xmax=60,
			axis y line*=right,
			yticklabels={,,},
			ytick style={draw=none},
			xticklabels={,,},
			xtick style={draw=none},
 			height=3.5cm,
		]
		\addplot[thick, densely dotted,color=cycle1] table[x=samples,y=acc] {data/exp/manifold-stability/training-evolution-cifar10-vgg16.dat};
		\addplot[name path=cifar10_top,color=cycle1!70,densely dotted,forget plot] table[x=samples,y=acc_cih] {data/exp/manifold-stability/training-evolution-cifar10-vgg16.dat};
		\addplot[name path=cifar10_btm,color=cycle1!70,densely dotted,forget plot] table[x=samples,y=acc_cil] {data/exp/manifold-stability/training-evolution-cifar10-vgg16.dat};
		\addplot[cycle1!50,fill opacity=0.5,forget plot] fill between[of=cifar10_top and cifar10_btm];
  \end{groupplot}
  \begin{groupplot}[group style={
                      group name=myplot,
                      group size= 2 by 1, horizontal sep=0.75cm},height=5cm, width=0.45\linewidth]
        \nextgroupplot[group/empty plot];
        \nextgroupplot[
			scale only axis,
			ylabel={Accuracy},
			xmin=1,
			xmax=60,
			axis lines=none,
			yticklabels={,,},
			ytick style={draw=none},
			xticklabels={,,},
			xtick style={draw=none},
 			height=3.5cm,
 			axis line style={draw=none},
		]	
		\addplot[thick, densely dotted,color=cycle2] table[x=samples,y=acc] {data/exp/manifold-stability/training-evolution-cifar100-vgg16.dat};
		\addplot[name path=cifar100_top,color=cycle2!70,densely dotted,forget plot] table[x=samples,y=acc_cih] {data/exp/manifold-stability/training-evolution-cifar100-vgg16.dat};
		\addplot[name path=cifar100_btm,color=cycle2!70,densely dotted,forget plot] table[x=samples,y=acc_cil] {data/exp/manifold-stability/training-evolution-cifar100-vgg16.dat};
		\addplot[cycle2!50,fill opacity=0.5,forget plot] fill between[of=cifar100_top and cifar100_btm];
  \end{groupplot}
\end{tikzpicture}
\vspace*{-3mm}
\caption{\textit{(left)} \thiswork{} score across convolutional layers of the VGG-16 network on CIFAR-10 and CIFAR-100 datasets; \textit{(right)} training progression in terms of accuracy (dotted) and \thiswork{} (solid) on CIFAR-10 and CIFAR-100 datasets for VGG-16 and ResNet-20, with respect to VGG-16.}\label{fig:nnmanifolds}
\end{figure*}

\subsection{Tracking the Evolution of Image Manifolds}\label{ssec:nn-internals}

Next, \dm{we employ \thiswork{} to inspect the internal dynamics of neural networks.}
\dm{We investigate \uline{the stability of output layer manifolds} across random initializations. We train~10 instances of the VGG-16~\citep{simonyan2015} network using different weight initializations on the CIFAR-10 and CIFAR-100 datasets. We compare the average \thiswork{} scores across representations in each network layer relative to the last layer. As Figure~\ref{fig:nnmanifolds} (left) shows, for both CIFAR-10 and CIFAR-100, the convolutional layers exhibit similar behavior; 
\thiswork{} shows that consequent layers do not monotonically contribute to the separation of image representations, but start to do so after initial feature extraction stage comprised of 4 convolutional blocks.
A low variance across the 10 networks trained from different random initializations indicates stability in the network structure.} %

We now \dm{\uline{examine the last network layers} during training with different initializations. Figure~\ref{fig:nnmanifolds} (right) plots the VGG16 validation errors and \thiswork{} scores relative to the final layer representations of \emph{two} pretrained networks, VGG16 itself with last layer dimension $d=512$ and ResNet-20 with $d=64$ and {\tt$\sim$}50 times less parameters. We observe that even in such unaligned spaces, \thiswork correctly identifies the convergence point of the networks. Surprisingly, we find that, in terms of \thiswork{}, VGG-16 representations progress towards not only the VGG-16 final layer, but the ResNet-20 final layer representation as well; this result suggests that these networks of distinct architectures share similar final structures.}

\begin{wrapfigure}[16]{r}{6cm}
\vspace{-5mm}
	\begin{tikzpicture}
	\begin{axis}[
		ylabel={Metric relative to MNIST},
		xlabel={Gaussian blur level $\sigma$},
		xmin=0,
		xmax=3,
		ymin=0,
		ymax=2,
		legend cell align=left,
		legend pos = north west,
		every axis x label/.style={
		    at={(ticklabel* cs:0.5)},
		    anchor=south,
		    below=8mm
		},
 		legend entries={\thiswork, FID, KID},
 		height=5cm,
 		width=6cm,
	]
	\def\imglevel{-0.35}
	\addlegendimage{line legend with two lines={thick, densely dotted, draw=cycle1}{ultra thick,color=cycle1,mark=*,mark size=1pt}};
	\addlegendimage{line legend with two lines={thick, dashed, draw=cycle2}{ultra thick,color=cycle2,mark=square*,mark size=1pt}};
	\addlegendimage{line legend with two lines={thick, loosely dashed, draw=cycle3}{ultra thick,color=cycle3,mark=triangle*,mark size=1pt}};
	
	\addplot[ultra thick,color=cycle1,mark=*,mark size=1pt] table[x=noise,y=lsd] {data/exp/sanity/noise/cifar10_100_blur_lsd.dat};
	\addplot[ultra thick,color=cycle2,mark=square*,mark size=1pt] table[x=noise,y=fid] {data/exp/sanity/noise/cifar10_100_blur_fid.dat};
	\addplot[ultra thick,color=cycle3,mark=triangle*,mark size=1pt] table[x=noise,y=kid] {data/exp/sanity/noise/cifar10_100_blur_kid.dat};

  	\draw [thick, loosely dashed, draw=cycle3] (axis cs: 0, 0.06062547527073234) -- (axis cs: 3, 0.06062547527073234);
  	\draw [thick, dashed, draw=cycle2] (axis cs: 0, 0.1987008072670359) -- (axis cs: 3, 0.1987008072670359);
  	\draw [thick, densely dotted, draw=cycle1] (axis cs: 0, 0.40533806169876085) -- (axis cs: 3, 0.40533806169876085);
  	\node[inner sep=0pt] (mnist_text) at (0.5, 0.40533806169876085+0.1) {\small CIFAR100};
  	
  	\draw [thick, solid, draw=black] (axis cs: 0, 1) -- (axis cs: 3, 1);
  	\node[inner sep=0pt] (mnist_text) at (0.5, 1+0.1) {\small MNIST};
  	
	\node[inner sep=0pt] (horse_0) at (0, \imglevel) {};
	\node[inner sep=0pt] (horse_10) at (1, \imglevel) {};
	\node[inner sep=0pt] (horse_20) at (2, \imglevel) {};
	\node[inner sep=0pt] (horse_30) at (3, \imglevel) {};
	\end{axis}
	\node[inner sep=0pt] (horse_0_over) at (horse_0) {\includegraphics[width=0.4cm]{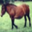}};
	\node[inner sep=0pt] (horse_10_over) at (horse_10) {\includegraphics[width=0.4cm]{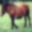}};
	\node[inner sep=0pt] (horse_20_over) at (horse_20) {\includegraphics[width=0.4cm]{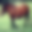}};
	\node[inner sep=0pt] (horse_30_over) at (horse_30) {\includegraphics[width=0.4cm]{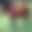}};
\end{tikzpicture}
	\caption{FID, KID and \thiswork{} on the \mbox{CIFAR-10} dataset with Gaussian blur.}\label{fig:blur}
\end{wrapfigure}

\vspace{-3mm}
\subsection{
Evaluating Generative Models}\label{ssec:gan-evaluation}

We now move on to apply \thiswork{} to evaluation of generative models.
First, \ants{we evaluate the \uline{sensitivity} of \thiswork{}, FID, and KID to simple image transformations as a proxy to more intricate artifacts of modern generative models.}
We \dm{progressively blur images from the CIFAR-10 training set, and measure the distance to the original data manifold, averaging outcomes over $100$ subsamples of $10$k images each. To enable comparison across methods, we normalize each distance measure such that the distance between CIFAR-10 and MNIST is $1$. 
Figure~\ref{fig:blur} reports the results at different levels $\sigma$ of Gaussian blur. 
We additionally report the normalized distance to the CIFAR-100 training set (dashed lines}
\begin{tikzpicture}
\draw[thick, densely dotted, draw=cycle1] plot coordinates {(0cm, -0.1cm) (0.75cm, -0.1cm)};
\draw[thick, dashed, draw=cycle2] plot coordinates {(0cm, 0cm) (0.75cm, 0cm)};
\draw[thick, loosely dashed, draw=cycle3] plot coordinates {(0cm, 0.1cm) (0.75cm, 0.1cm)};
\end{tikzpicture}).
\dm{
FID and KID quickly drift away from the original distribution and match MNIST, a dataset of a completely different nature. \ants{Contrariwise, \thiswork{} is more robust to noise and follows the datasets structure, as the relationships between objects remain mostly unaffected on low blur levels.} Moreover, with both FID and KID, low noise ($\sigma=1$) applied to CIFAR-10 suffices to exceed the distance of CIFAR-100, which is similar to CIFAR-10. 
\thiswork{} is much more robust, exceeding that distance only with $\sigma=2$.}

\begin{table}[!t]
\begin{center}
\small
\setlength{\tabcolsep}{1pt}
\renewcommand{\aboverulesep}{0pt}
\renewcommand{\belowrulesep}{0pt}
\newcolumntype{C}{>{\centering\arraybackslash}X}
\caption{\thiswork agrees with KID and FID across varying datasets for GAN evaluation.}\label{tab:gans}
\begin{tabularx}{\textwidth}{p{1.2cm}CCCCCCCC}
\toprule
&
\multicolumn{2}{c}{\textbf{MNIST}} & 
\multicolumn{2}{c}{\textbf{FashionMNIST}} &
\multicolumn{2}{c}{\textbf{CIFAR10}} &
\multicolumn{2}{c}{\textbf{CelebA}} \\
\cmidrule(lr){2-3}\cmidrule(lr){4-5}\cmidrule(lr){6-7}\cmidrule(lr){8-9}
\textbf{Metric}    & WGAN & WGAN-GP & WGAN & WGAN-GP & WGAN & WGAN-GP & WGAN & WGAN-GP \\ 
\midrule
\mbox{\thiswork} 
      & $\mathbf{57.74} \pm 0.47 $\phantom{$\pm$}
	  & $\mathbf{10.77} \pm 0.42 $\phantom{$\pm$}
      & $\mathbf{118.14} \pm 0.52 $\phantom{$\pm$}
      & $\mathbf{13.45} \pm 0.54 $\phantom{$\pm$}
      & $\mathbf{18.10} \pm 0.36 $\phantom{$\pm$}
      & $\mathbf{10.84} \pm 0.42 $\phantom{$\pm$}
      & $\mathbf{10.11} \pm 0.33 $\phantom{$\pm$}
      & $\mathbf{2.84} \pm 0.31 $\phantom{$\pm$} \\
\mbox{KID $\times 10^3$}
       & $\mathbf{47.26} \pm 0.07 $\phantom{$\pm$}
       & $\mathbf{ 5.53} \pm 0.03 $\phantom{$\pm$}
       & $\mathbf{119.93} \pm 0.14 $\phantom{$\pm$}
       & $\mathbf{ 25.49} \pm 0.07 $\phantom{$\pm$}
       & $\mathbf{ 93.89} \pm 0.09 $\phantom{$\pm$}
       & $\mathbf{ 59.59} \pm 0.09 $\phantom{$\pm$}
       & $\mathbf{217.28} \pm 0.14 $\phantom{$\pm$}
       & $\mathbf{ 92.71} \pm 0.08 $\phantom{$\pm$} \\
\mbox{FID} 
       & $\mathbf{31.75} \pm 0.07 $\phantom{$\pm$}
       & $\mathbf{8.95} \pm 0.03 $\phantom{$\pm$}
       & $\mathbf{152.44} \pm 0.12 $\phantom{$\pm$}
       & $\mathbf{35.31} \pm 0.07 $\phantom{$\pm$}
       & $\mathbf{101.43} \pm 0.09 $\phantom{$\pm$}
       & $\mathbf{80.65} \pm 0.09 $\phantom{$\pm$}
       & $\mathbf{205.63} \pm 0.09 $\phantom{$\pm$}
       & $\mathbf{85.55} \pm 0.08 $\phantom{$\pm$} \\
\bottomrule
\end{tabularx}
\end{center}
\vspace{-7mm}
\end{table} 
\begin{wrapfigure}[15]{r}{6cm}
\vspace{-5mm}
	\begin{tikzpicture}
	\begin{axis}[
		ylabel={$\hkt_G$ with ER normalization},
		xlabel={$t$},
		xmin=1,
		xmax=100,
		xmode=log,
		ymode=log,
		legend cell align=left,
		legend pos = north west,
        legend columns=2, 
        width=6cm,
        height=5cm,
        ytick style={draw=none}
	]
	\addplot[thick,color=cycle1] table[x=samples,y=mnist] {data/exp/interpretability.dat};
	\addlegendentry{MNIST};
	\addplot[thick,color=cycle2] table[x=samples,y=fmnist] {data/exp/interpretability.dat};
	\addlegendentry{FMNIST};
	\addplot[thick,color=cycle3] table[x=samples,y=cifar10] {data/exp/interpretability.dat};
	\addlegendentry{CIFAR10};
	\addplot[thick,color=cycle4] table[x=samples,y=cifar100] {data/exp/interpretability.dat};
	\addlegendentry{CIFAR100};
	\addplot[thick,color=cycle5] table[x=samples,y=celeba] {data/exp/interpretability.dat};
	\addlegendentry{CelebA};
	\end{axis}
\end{tikzpicture}
\vspace*{-9mm}
	\caption{Plotting the normalized heat trace allows interpretation of medium- and global-scale structure of datasets. Best viewed in color.}\label{fig:interpretability}
\end{wrapfigure}
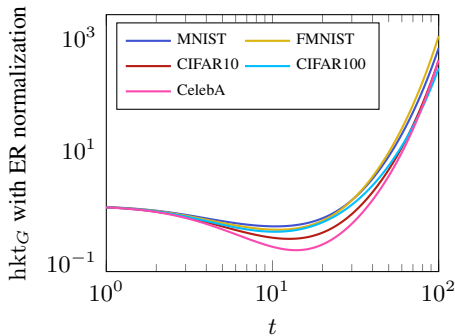
 Next, we turn our attention to the \uline{sample-based evaluation} of generative models.
We then train the WGAN~\citep{arjovsky2017} and WGAN-GP~\citep{gulrajani2017} models on four datasets: MNIST, FashionMNIST, CIFAR10 and CelebA. We sample $10$k samples, $\mY$, from each GAN. We then uniformly subsample $10k$ images from the corresponding original dataset, $\mX$, and compute the \thiswork{}, KID and FID scores between $\mX$ and $\mY$. Table~\ref{tab:gans} reports the average measure and its 99\% confidence interval across $100$ runs. \thiswork, as well as both FID and KID, reflect the fact that WGAN-GP is a more expressive model. We provide details on architecture, training, and generated samples in Appendix C.
\ants{Additionally, in Appendix~\ref{ssec:extra-experiments} we demonstrate superiority of \thiswork on synthetic data.}

\subsection{Interpreting \thiswork{}}\label{ssec:interpretability}
\vspace{-2mm}
To \ants{understand how \thiswork{} operates, we investigate the behavior of heat kernel traces of different datasets that are normalized by a null model. \citet{tsitsulin2018} proposed a normalization by the heat kernel trace of an empty graph, which amounts to taking the average, rather than the sum, of the original heat kernel diagonal. However, this normalization is not an appropriate null model as it ignores graph connectivity. We propose a heat kernel normalization by the \emph{expected} heat kernel of an Erd{\H{o}}s\hyp{}R{\'e}nyi graph (further details in the Appendix~\ref{ssec:expset}}).

\ants{Figure~\ref{fig:interpretability} depicts the obtained normalized $\hkt_g$ for all datasets we work with.
We average results over $100$ subsamples of $10k$ images each.
For $t=10$, i.e., at a medium scale, CelebA is most different from the random graph,
while for large-scale $t$ values, which capture global community structure, $ %
\frac{\mathrm{d}\hkt_g(t)}{\mathrm{d}t}$ reflects the approximate number of clusters in the data. Surprisingly, CIFAR-100 comes close to CIFAR-10 for large $t$ values; we have found that this is due to the fact that the pre-trained Inception network does not separate the CIFAR-100 data classes well enough.}
We conclude that the heat kernel trace is interpretable if we normalize it with an appropriate null model.
\begin{figure*}[!t]
	\begin{tikzpicture}
    \begin{groupplot}[group style={
                      group name=myplot,
                      group size= 2 by 1, horizontal sep=1.5cm},height=4.5cm]
        \nextgroupplot[
			ylabel={Metric relative to $10^4$ samples},
			xlabel={Number of samples},
			xmin=900,
			xmax=50000,
			xmode=log,
			ymin=0.5,
			ymax=2.5,
			extra x ticks={50000},
			width=0.5\linewidth
		]
		\addplot[ultra thick,color=cycle1,mark=*,mark size=1pt] table[x=samples,y=lsd] {data/exp/stability/cifar10_100_samples_lsd.dat};
		\addplot[name path=lsd_top,color=cycle1!70,forget plot] table[x=samples,y=cih] {data/exp/stability/cifar10_100_samples_lsd.dat};
		\addplot[name path=lsd_btm,color=cycle1!70,forget plot] table[x=samples,y=cil] {data/exp/stability/cifar10_100_samples_lsd.dat};
		\addplot[cycle1!50,fill opacity=0.5,forget plot] fill between[of=lsd_top and lsd_btm];
		\addplot[ultra thick,color=cycle6,mark=diamond*,mark size=1pt] table[x=samples,y=lsd_kgraph_max_empty] {data/exp/stability/cifar10_100_samples_lsd_kgraph_max_empty_.dat};
		\addplot[name path=lsd_top_,color=cycle6!70,forget plot] table[x=samples,y=cih] {data/exp/stability/cifar10_100_samples_lsd_kgraph_max_empty_.dat};
		\addplot[name path=lsd_btm_,color=cycle6!70,forget plot] table[x=samples,y=cil] {data/exp/stability/cifar10_100_samples_lsd_kgraph_max_empty_.dat};
		\addplot[cycle6!50,fill opacity=0.5,forget plot] fill between[of=lsd_top_ and lsd_btm_];
		\addplot[ultra thick,color=cycle2,mark=square*,mark size=1pt] table[x=samples,y=fid] {data/exp/stability/cifar10_100_samples_fid.dat};
		\addplot[name path=fid_top,color=cycle2!70,forget plot] table[x=samples,y=cih] {data/exp/stability/cifar10_100_samples_fid.dat};
		\addplot[name path=fid_btm,color=cycle2!70,forget plot] table[x=samples,y=cil] {data/exp/stability/cifar10_100_samples_fid.dat};
		\addplot[cycle2!50,fill opacity=0.5,forget plot] fill between[of=fid_top and fid_btm];
		\addplot[ultra thick,color=cycle3,mark=triangle*,mark size=1pt] table[x=samples,y=kid] {data/exp/stability/cifar10_100_samples_kid.dat};
		\addplot[name path=kid_top,color=cycle3!70,forget plot] table[x=samples,y=cih] {data/exp/stability/cifar10_100_samples_kid.dat};
		\addplot[name path=kid_btm,color=cycle3!70,forget plot] table[x=samples,y=cil] {data/exp/stability/cifar10_100_samples_kid.dat};
		\addplot[cycle3!50,fill opacity=0.5,forget plot] fill between[of=kid_top and kid_btm];
		\nextgroupplot[
			ylabel={Execution time, seconds},
			xlabel={Number of samples},
			xmin=3000,
			xmode=log,
			ymode=log,
			legend cell align=left,
			legend pos = north east,
     	   legend columns=2, 
			xtick={3162.27766017, 10000},
			extra x ticks={50000},
			width=0.5\linewidth
		]
		\addplot[ultra thick,color=cycle1,mark=*,mark size=1pt] table[x=samples,y=lsd] {data/exp/scalability/nsamples_d_2048_lsd.dat};
		\addplot[name path=lsd_top,color=cycle1!70,forget plot] table[x=samples,y=cih] {data/exp/scalability/nsamples_d_2048_lsd.dat};
		\addplot[name path=lsd_btm,color=cycle1!70,forget plot] table[x=samples,y=cil] {data/exp/scalability/nsamples_d_2048_lsd.dat};
		\addplot[cycle1!50,fill opacity=0.5,forget plot] fill between[of=lsd_top and lsd_btm];
		\addplot[ultra thick,color=cycle6,mark=diamond*,mark size=1pt] table[x=samples,y=lsd_kg] {data/exp/scalability/nsamples_d_2048_lsd_kg.dat};
		\addplot[name path=lsd_kg_top,color=cycle6!70,forget plot] table[x=samples,y=cih] {data/exp/scalability/nsamples_d_2048_lsd_kg.dat};
		\addplot[name path=lsd_kg_btm,color=cycle6!70,forget plot] table[x=samples,y=cil] {data/exp/scalability/nsamples_d_2048_lsd_kg.dat};
		\addplot[cycle6!50,fill opacity=0.5,forget plot] fill between[of=lsd_kg_top and lsd_kg_btm];
		\addplot[ultra thick,color=cycle2,mark=square*,mark size=1pt] table[x=samples,y=fid] {data/exp/scalability/nsamples_d_2048_fid.dat};
		\addplot[name path=fid_top,color=cycle2!70,forget plot] table[x=samples,y=cih] {data/exp/scalability/nsamples_d_2048_fid.dat};
		\addplot[name path=fid_btm,color=cycle2!70,forget plot] table[x=samples,y=cil] {data/exp/scalability/nsamples_d_2048_fid.dat};
		\addplot[cycle2!50,fill opacity=0.5,forget plot] fill between[of=fid_top and fid_btm];
		\addplot[ultra thick,color=cycle3,mark=triangle*,mark size=1pt] table[x=samples,y=kid] {data/exp/scalability/nsamples_d_2048_kid.dat};
		\addplot[name path=kid_top,color=cycle3!70,forget plot] table[x=samples,y=cih] {data/exp/scalability/nsamples_d_2048_kid.dat};
		\addplot[name path=kid_btm,color=cycle3!70,forget plot] table[x=samples,y=cil] {data/exp/scalability/nsamples_d_2048_kid.dat};
		\addplot[cycle3!50,fill opacity=0.5,forget plot] fill between[of=kid_top and kid_btm];
		\addplot[ultra thick,color=cycle4,mark=pentagon*,mark size=1pt] table[x=samples,y=gs] {data/exp/scalability/nsamples_d_2048_gs.dat};
		\addplot[name path=gs_top,color=cycle4!70,forget plot] table[x=samples,y=cih] {data/exp/scalability/nsamples_d_2048_gs.dat};
		\addplot[name path=gs_btm,color=cycle4!70,forget plot] table[x=samples,y=cil] {data/exp/scalability/nsamples_d_2048_gs.dat};
		\addplot[cycle4!50,fill opacity=0.5,forget plot] fill between[of=gs_top and gs_btm];
  \end{groupplot}
\path (myplot c1r1.north west|-current bounding box.north)--
      coordinate(legendpos)
      (myplot c2r1.north east|-current bounding box.north);
\matrix[
    matrix of nodes,
    anchor=south,
    draw,
    inner sep=0.2em,
    draw
  ]at([yshift=1ex]legendpos)
  {
    \draw[mark phase=2, mark repeat=2, very thick,color=cycle1,mark=*,mark size=1pt] plot coordinates {(0cm, 0.1cm) (0.25cm, 0.1cm) (0.5cm, 0.1cm)}; & \thiswork (exact) & [5pt]
    \draw[mark phase=2, mark repeat=2, very thick,color=cycle6,mark=diamond*,mark size=1pt] plot coordinates {(0cm, 0.1cm) (0.25cm, 0.1cm) (0.5cm, 0.1cm)}; & \thiswork (approx.) & [5pt]
    \draw[mark phase=2, mark repeat=2, very thick,color=cycle2,mark=square*,mark size=1pt] plot coordinates {(0cm, 0.1cm) (0.25cm, 0.1cm) (0.5cm, 0.1cm)}; & FID & [5pt]
    \draw[mark phase=2, mark repeat=2, very thick,color=cycle3,mark=triangle*,mark size=1pt] plot coordinates {(0cm, 0.1cm) (0.25cm, 0.1cm) (0.5cm, 0.1cm)}; & KID & [5pt]
    \draw[mark phase=2, mark repeat=2, very thick,color=cycle4,mark=pentagon*,mark size=1pt] plot coordinates {(0cm, 0.1cm) (0.25cm, 0.1cm) (0.5cm, 0.1cm)}; & GS & [5pt]\\
  };
\end{tikzpicture}
\vspace*{-5mm}
\caption{Stability and scalability experiment: \textit{(left)} stability of FID, KID and \thiswork{} wrt.\ sample size on CIFAR-10 and CIFAR-100 dataset; \textit{(right)} scalability of FID, KID and \thiswork{} wrt.\ sample size on synthetic datasets.}\label{fig:scalability}
\vspace{-5mm}
\end{figure*}
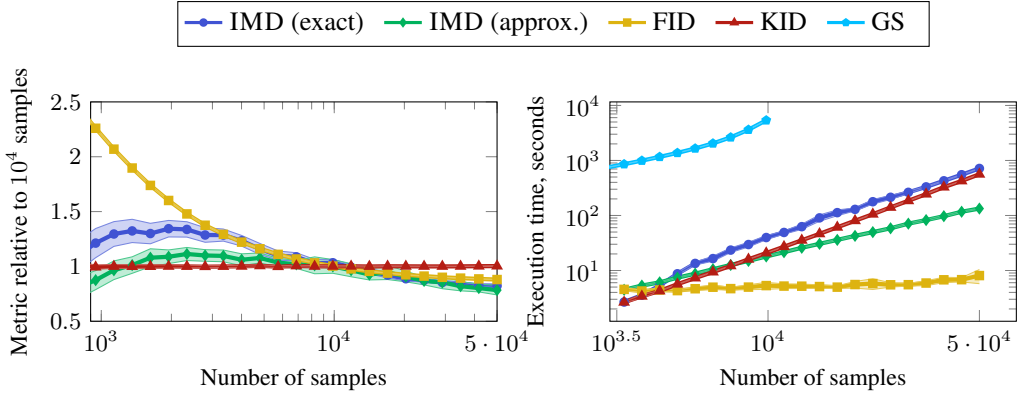 
\subsection{Verifying Stability and Scalability of \thiswork{}}\label{ssec:stability-scalability}
\vspace{-2mm}
\dm{In addition to the complexity analysis in Section~\ref{ssec:thiswork}, we assess the \uline{scaling and sample stability} of \thiswork{}. Since \thiswork{}, like FID, is a lower bound to an optimal transport-based metric, we cannot hope for an unbiased estimator. However, we empirically verify, in Figure~\ref{fig:scalability} (left), that \thiswork{} does not diverge too much with increased sample size. Most remarkably, we observe that \thiswork{} with approximate $\knn$~\citep{dong2011} does not induce additional variance, while it diverges slightly further than the exact version as the number of samples grows. %

In terms of \uline{scalability}, Figure~\ref{fig:scalability} (right) shows that the theoretical complexity is supported in practice. Using approximate $\knn$, we break the $\bigO(n^2)$ performance of KID. FID's time complexity appears constant, as its runtime is dominated by the $\bigO(d^3)$ matrix square root operation.
Geometry score (GS) fails to perform scalably, as its runtime grows exponentially. Due to this prohibitive computational cost, we eschew other comparison with GS. Furthermore, as \thiswork{} distance is computed through a low-dimensional heat trace representation of the manifold, we can store HKT for future comparisons, thereby enhancing performance in the case of many-to-many comparisons.}
\vspace{-2mm} %
\section{Discussion and Future Work}\label{sec:conclusions}
\vspace{-2mm}
\dm{We introduced \thiswork{}, \pki{a geometry-grounded, first-of-its-kind} intrinsic multi-scale method for comparing unaligned manifolds,}
\pki{which we approximate efficiently with guarantees, utilizing the Stochastic Lanczos Quadrature.}
\pki{We have shown the expressiveness of \thiswork{} in quantifying the change of data representations in NLP and image processing, evaluating generative models, and in the study of neural network representations.}
\pki{Since \thiswork{} allows comparing diverse manifolds,
its applicability is not limited to the tasks we have evaluated, while it
paves the way to the development of even more expressive techniques founded on geometric insights.} %
\section*{Acknowledgements}
This work was partially funded by the Ministry of Science and Education of Russian Federation as a part of Mega Grant Research Project 14.756.31.0001. Ivan Oseledets would like to thank Huawei for the support of his research. {
\bibliographystyle{iclr2020_conference}
\bibliography{main}
}
\newpage
\section*{Appendix}
\addcontentsline{toc}{section}{Appendices}
\renewcommand{\thesubsection}{\Alph{subsection}}
\subsection{Trace estimation error bounds}
\label{ssec:slqerrs}
We will use the error of the Lanczos approximation of the action of the matrix exponential ${f(\mL)\vV = \exp^{-t \mL} \vV}$ to estimate the error of the trace. We first rewrite quadratic form under summation in the trace approximation to a convenient form,
\begin{equation}
\label{eq:quadform}
	\vV^\top f(\mL) \vV \approx \sum\limits_{k=0}^m \tau_k^2 f(\lambda_k) = \sum_{k=0}^m [\vE_1^\top \vU_k]^2 f(\lambda_k) = \vE_1^\top \mU f(\Lambda) \mU^\top \vE_1 = \vE_1^\top f(\mT) \vE_1.
\end{equation}
Because the Krylov subspace $\mathcal{K}_m(\mL, \vV)$ is built on top of vector $\vV$ with $\mQ$ as an orthogonal basis of $\mathcal{K}_m(\mL, \vV)$, i.e. $\vQ_0 = \vV$ and $\vV \perp \vQ_i$ for $i \in (1, \dots, m-1)$, the following holds
\begin{equation}
	\vV^\top f(\mL) \vV \approx \vV^\top \mQ f(\mT)\vE_1 =\vE_1^\top f(\mT)\vE_1.
\end{equation}
Thus, the error in quadratic form estimate $\vV^\top f(\mL) \vV$ is exactly the error of Lanczos approximation $f(\mL)\vV \approx \mQ f(\mT) \vE_1$. 
To obtain the error bounds, we use the Theorem 2 in~\citet{hochbruck1997}, which we recite below.
\begin{theorem}\label{thm:hochbruck-arnoldierror}
Let $\mL$ be a real symmetric positive semi-definite matrix with eigenvalues in the interval $[0, 4\rho]$. Then the error in the $m$-step Lanczos approximation of ${\exp^{-t \mL}\vV}$, i.e. ${\epsilon_m = \| \exp^{-t \mL}\vV - \mQ_m \exp^{-t \mT_m}\vE_1 \|}$, is bounded in the following ways:
\begin{subnumcases}{ \epsilon_m \leq}
	{10 e^{-m^2/(5\rho t)}}, & ${\sqrt{4\rho t} \leq m \leq 2\rho t}$ \\
	{10 (\rho t)^{-1} e^{-\rho t} \Big( \frac{e\rho t}{m}\Big)^m}, & ${m \geq 2\rho t}$
\end{subnumcases}
\end{theorem}

Since $\vV$ is a unit vector, thanks to Cauchy–Bunyakovsky–Schwarz inequality, we can upper\hyp{}bound the error of the quadratic form approximation by the error of the ${\exp^{-t \mL}\vV}$ approximation, i.e. ${|\vV^\top f(\mL) \vV - \vE_1^\top \mU f(\Lambda) \mU^\top \vE_1| \leq \| \exp^{- t \mL}\vV - \mQ_m \exp^{- t \mT_m}\vE_1 \| = \epsilon_m}$.

Following the argumentation in \citet{ubaru2017}, we obtain a condition on the number of Lanczos steps $m$ by setting
${\epsilon_m \leq \frac{\epsilon}{2} f_{min}(\lambda)}$,
where $f_{min}(\lambda)$ is the minimum value of $f$ on $[\lambda_{min}, \lambda_{max}]$.
We now derive the absolute error between the Hutchinson estimate of Equation~(\textcolor{deepblue}{3}) and the SLQ of Equation~(\textcolor{deepblue}{6}):
\begin{align*}
	\Big|\tr_{n_v}(f(\mL)) - \Gamma \Big| &= \frac{n}{n_v} \Bigg | \sum_{i=1}^{n_v} \vV_i^\top f(\mL) \vV_i - \sum_{i=1}^{n_v} \vE_1^\top f(\mT^{(i)}) \vE_1 \Bigg|  \\
	&\leq \frac{n}{n_v} \sum_{i=1}^{n_v} \Bigg | \vV_i^\top f(\mL) \vV_i -\vE_1^\top f(\mT^{(i)})\vE_1 \Bigg| \\
	&\leq \frac{n}{n_v} \sum_{i=1}^{n_v} \epsilon_m = n\epsilon_m ,
\end{align*}
where $\mT^{(i)}$ is the tridiagonal matrix obtained with Lanczos algorithm with starting vector $\vV_i$.
\begin{equation}
	\Big|\tr_{n_v} f(\mL) - \Gamma \Big| \leq n\epsilon_m \leq \frac{n\epsilon}{2}f_{min}(\lambda) \leq \frac{\epsilon}{2}\tr(f(\mL)),
\end{equation}
Finally, we formulate SLQ as an $(\epsilon, \delta)$ estimator,
\begin{align*}
	1 - \delta &\leq \pr\Bigg[\Big| \tr(f(\mL)) - \tr_{n_v}(f(\mL))\Big| \leq \frac{\epsilon}{2} \Big|\tr(f(\mL)) \Big|\Bigg] \\
	&\leq \pr\Bigg[\Big| \tr(f(\mL)) -  \tr_{n_v}(f(\mL)) \Big| +
	\Big| \tr_{n_v}(f(\mL)) - \Gamma \Big|	
	\leq \frac{\epsilon}{2}\Big|\tr(f(\mL))\Big| + \frac{\epsilon}{2}\Big|\tr(f(\mL))\Big| \Bigg] \\
	&\leq  \pr\Bigg[\Big| \tr(f(\mL)) - \Gamma\Big|\leq \epsilon \Big|\tr(f(\mL))\Big|\Bigg],
\end{align*}

For the normalized Laplacian $\mL$, the minimum eigenvalue is $0$ and $f_{\min}(0) = \exp(0) = 1$, hence $\epsilon_m \leq \frac{\epsilon}{2}$, and the eigenvalue interval has $\rho = 0.5$. We can thus derive the appropriate number of Lanczos steps $m$ to achieve error $\epsilon$,
\begin{subnumcases}{\label{eq:eps2} \epsilon \leq }
{20 e^{-m^2/(2.5 t)}}, & $\sqrt{2t} \leq m \leq t$ \label{eq:eps2a}\\
{40t^{-1} e^{-0.5t} \Big( \frac{0.5et}{m}\Big)^m}, & $ m \geq t$ \label{eq:eps2b}
\end{subnumcases}

\begin{wrapfigure}[21]{r}{6cm}
\vspace{-5mm}
	\begin{tikzpicture}
	\begin{axis}[
		ylabel={Matvec approximation error, $\epsilon_m$},
		xlabel={Number of Lanczos steps, $m$},
		xmin=1,
		xmax=21,
		ymode=log,
		ymin=1e-19,
		ymax=1,
		legend cell align=left,
		legend style={at={(0.375,1.05)},anchor=south},
        legend columns=3, 
        extra x ticks=1,
        width=6cm,
        height=6cm
	]
	\addplot[very thick,color=cycle1,mark=*,mark size=1pt] table {data/errs/mverrs_t=0.1.dat};
	\addlegendentry{$t=0.1$}
	
	\addplot[very thick,color=cycle3,mark=triangle*,mark size=1pt] table {data/errs/mverrs_t=5.05.dat};
	\addlegendentry{$t=5.05$}
	
	\addplot[very thick,color=cycle5,mark=square*,mark size=1pt] table {data/errs/mverrs_t=10.0.dat};
	\addlegendentry{$t=10.0$}
	
	\addplot[very thick,color=cycle2,mark=x,mark size=2pt] table {data/errs/mverrs_t=2.575.dat};
	\addlegendentry{$t=2.575$}
	
	\addplot[very thick,color=cycle4,mark=diamond*,mark size=1pt] table {data/errs/mverrs_t=7.525.dat};
	\addlegendentry{$t=7.525$}

	\addplot[dotted, ultra thick,color=cycle1] table {data/errs/mverrbs_t=0.1.dat};
	\addplot[dotted, ultra thick,color=cycle2] table {data/errs/mverrbs_t=2.575.dat};
	\addplot[dotted, ultra thick,color=cycle3] table {data/errs/mverrbs_t=5.05.dat};
	\addplot[dotted, ultra thick,color=cycle4] table {data/errs/mverrbs_t=7.525.dat};
	\addplot[dotted, ultra thick,color=cycle5] table {data/errs/mverrbs_t=10.0.dat};
	\end{axis}
\end{tikzpicture}
	\caption{Errors (solid) and error bounds (dotted) for the approximation of matrix exponential action with varying temperature $t$.}
	\label{fig:matvec-errors}
\end{wrapfigure}
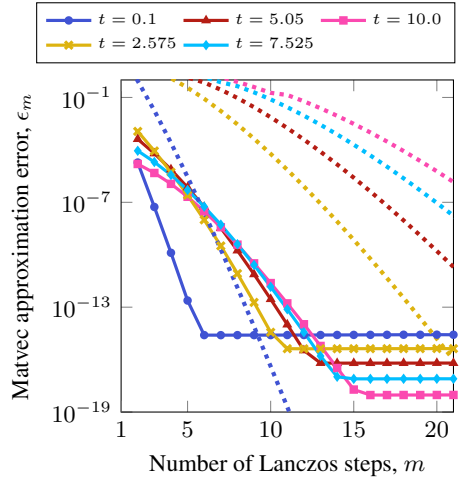
 
\begin{figure*}[!th]
	\begin{tikzpicture}
    \begin{groupplot}[group style={
                      group name=myplot,
                      group size= 2 by 1, horizontal sep=0.2cm},height=6cm,width=0.499\linewidth]
        \nextgroupplot[
		ylabel={Relative trace approximation error, $\epsilon$},xlabel={Number of Lanczos steps, $m$},ymode=log,extra x ticks=1, ymin=1e-9, ymax=1]
					\addplot[very thick,color=cycle1,mark=*,mark size=1pt] table {data/errs/slqerr_t=0.1.dat};
					\addplot[very thick,color=cycle2,mark=triangle*,mark size=1pt] table {data/errs/slqerr_t=1.0.dat};
					\addplot[very thick,color=cycle3,mark=square*,mark size=1pt] table {data/errs/slqerr_t=2.0.dat};
					\addplot[very thick,color=cycle4,mark=x,mark size=2pt] table {data/errs/slqerr_t=10.0.dat};
	
					\addplot[dotted, ultra thick,color=cycle1] table {data/errs/trerrs_t=0.1.dat};
					\addplot[dotted, ultra thick,color=cycle2] table {data/errs/trerrs_t=1.0.dat};
					\addplot[dotted, ultra thick,color=cycle3] table {data/errs/trerrs_t=2.0.dat};
					\addplot[dotted, ultra thick,color=cycle4] table {data/errs/trerrs_t=10.0.dat};    
		\nextgroupplot[xlabel={Number of random vectors, $n_v$}, xmode=log, ymode=log, ymin=1e-9, ymax=1,yticklabels={,,},]
					\addplot[very thick,color=cycle1,mark=*,mark size=1pt] table {data/errs/nvslqerr_t=0.1.dat};\label{lines:t-01};
					\addplot[very thick,color=cycle2,mark=triangle*,mark size=1pt] table {data/errs/nvslqerr_t=1.0.dat};\label{lines:t-1};
					\addplot[very thick,color=cycle3,mark=square*,mark size=1pt] table {data/errs/nvslqerr_t=2.0.dat};\label{lines:t-2};
					\addplot[very thick,color=cycle4,mark=x,mark size=2pt] table {data/errs/nvslqerr_t=10.0.dat};\label{lines:t-10};
					\addplot[dotted, ultra thick,color=black] table {data/errs/nveps.dat};\label{lines:theory-nv};
  \end{groupplot}
\path (myplot c1r1.north west|-current bounding box.north)--
      coordinate(legendpos)
      (myplot c2r1.north east|-current bounding box.north);
\matrix[
    matrix of nodes,
    anchor=south,
    draw,
    inner sep=0.2em,
    draw
  ]at([yshift=1ex]legendpos)
  {
    \draw[mark phase=2, mark repeat=2, very thick,color=cycle1,mark=*,mark size=1pt] plot coordinates {(0cm, 0.1cm) (0.25cm, 0.1cm) (0.5cm, 0.1cm)}; & $t=0.1$ & [5pt]
    \draw[mark phase=2, mark repeat=2, very thick,color=cycle2,mark=*,mark size=1pt] plot coordinates {(0cm, 0.1cm) (0.25cm, 0.1cm) (0.5cm, 0.1cm)}; & $t=1$ & [5pt]
    \draw[mark phase=2, mark repeat=2, very thick,color=cycle3,mark=*,mark size=1pt] plot coordinates {(0cm, 0.1cm) (0.25cm, 0.1cm) (0.5cm, 0.1cm)}; & $t=2$ & [5pt]
    \draw[mark phase=2, mark repeat=2, very thick,color=cycle4,mark=*,mark size=1pt] plot coordinates {(0cm, 0.1cm) (0.25cm, 0.1cm) (0.5cm, 0.1cm)}; & $t=10$ & [5pt]
    \draw[dotted, ultra thick,color=black] plot coordinates {(0cm, 0.1cm) (0.25cm, 0.1cm) (0.5cm, 0.1cm)}; & bound & [5pt]\\
  };
\end{tikzpicture}
\caption{Trace estimation errors (solid) and error bounds (dotted) for: \textit{(left)} the number of Lanczos steps $m$ with fixed number of random vectors $n_v = 100$; \textit{(right)} the number of random vectors $n_v$ in Hutchinson estimator with fixed number of Lanczos steps $m = 10$. Lines correspond to varying temperatures $t$.} \label{fig:errors}
\end{figure*}
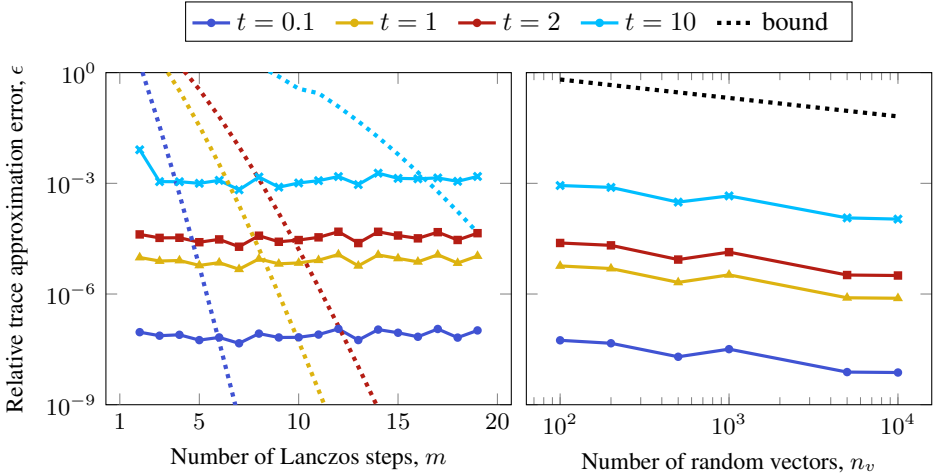

Figure~\ref{fig:matvec-errors} shows the tightness of the bound for the approximation of the matrix exponential action on vector $\vV$, $\epsilon_m = \| \exp(-t\mL) - \mQ_m \exp(-t\mT_m) \vE_1 \|$. We can see that for most of the temperatures $t$, very few Lanczos steps $m$ are sufficient, i.e. we can set $m = 10$. However, the error from the Hutchinson estimator dominates the overall error. Figure~\ref{fig:errors} shows the error of trace estimation does not change with $m$ and for $t=0.1$ is around $10^{-3}$. In case of a Rademacher $p(\vV)$, the bound on the number of random samples is ${n_v \geq \frac{6}{\epsilon^2} \log(2/\delta)}$~\citep{roosta2015}. Employing 10k vectors results in the error bound of roughly $10^{-2}$. In practice, we observe the performance much better than given by the bound, see Figure~\ref{fig:errors}.

One particular benefit of small $m$ value is that we do not have to worry about the orthogonality loss in the Lanczos algorithm which often undermines its convergence. Since we do only a few Lanczos iterations, the rounding errors hardly accumulate causing little burden in terms of orthogonality loss between the basis vectors of the Krylov subspace.

\subsection{Variance reduction}
We reduce variance of the randomized estimator through control variates. The idea is to use Taylor expansion to substitute a part of the trace estimate with its easily computed precise value,
\begin{align}
	\tr(\exp(-t\mL)) &= \texttt{slq}\Big[\exp(-t\mL) - (\mI -t\mL + \frac{t^2\mL^2}{2}) \Big] + \tr(\mI - t\mL + \frac{t^2\mL^2}{2}) \\
	&= \texttt{slq}\Big[\exp(-t\mL) - (\mI -t\mL + \frac{t^2\mL^2}{2}) \Big] + n + \tr(-t\mL) + \frac{t^2 \| \mL\|_F^2}{2} \\
	&= \texttt{slq}\Big[\exp(-t\mL) \Big] + \texttt{slq}\Big[t\mL\Big] - \texttt{slq}\Big[\frac{t^2\mL^2}{2})\Big] - tn + \frac{t^2 \| \mL\|_F^2}{2},
\end{align}
where we use the fact that $\| \mL \|_F = \sqrt{\tr(\mL^\top \mL)}$ and that the trace of normalized Laplacian is equal to $n$. It does reduce the variance of the trace estimate for smaller temperatures $t\leq 1$.

\begin{wrapfigure}[13]{r}{6cm}
\vspace{-5mm}
	\begin{tikzpicture}
	\begin{axis}[
		ylabel={Standard variation, $\sigma$},
		xlabel={Temperature, $t$},
		xmin=0.1,
		xmax=10,
		xmode=log,
		ymode=log,
		legend cell align=left,
		legend style={at={(0.5,0.05)},anchor=south},
		legend entries={original SLQ, SLQ w/ variance reduction},
		height = 5cm,
		width = 6cm
	]
	\addplot[very thick,color=cycle2,mark=*,mark size=1pt] table {data/errs/origslqtr.dat};
	\addplot[very thick,color=cycle1,mark=square*,mark size=1pt] table {data/errs/redvartr.dat};
	
	\end{axis}
\end{tikzpicture}
\vspace*{-6mm}
	\caption{Variance of the trace estimate.}\label{fig:variance}
\end{wrapfigure}
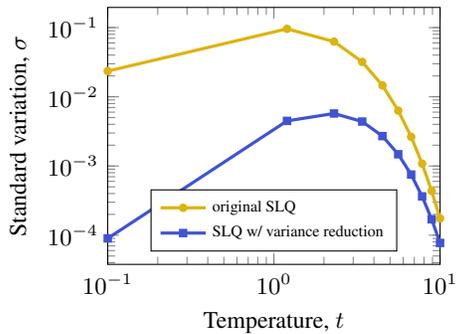

To obtain this advantage over the whole range of $t$, we utilize the following variance reduction form:
\begin{align}
	\tr(\exp(-t\mL)) &= \texttt{slq}\Big[\exp(-t\mL) - (\mI - \alpha t\mL) \Big ] +  n (1 - \alpha t),
\end{align}
where there exists an alpha that is optimal for every $t$, namely setting $\alpha = 1/\exp(t)$.
We can see the variance reduction that comes from this procedure in the Figure~\ref{fig:variance}.

\subsection{Experiments Discussion}\label{ssec:extra-experiments}

\ants{Here we include additional results that did not find their way to the main paper body.}
\subsubsection{FID and KID Fail to Find Structure in Unaligned Corpora}

\begin{figure*}[!t]
	\begin{tikzpicture}
    \begin{groupplot}[group style={
                      group name=myplot,
                      group size= 2 by 1, horizontal sep=2.5cm},width=.45\textwidth,height=.45\textwidth,
xticklabels={pl,ru,el,hu,tr,ar,he,en,simple,sv,de,es,nl,pt,vi,war},
yticklabels={pl,ru,el,hu,tr,ar,he,en,simple,sv,de,es,nl,pt,vi,war},
	xtick={0,...,15},
    ytick={15,...,0},
    ytick style={draw=none},
    xtick style={draw=none},
    xmin=0,
    xmax=15,
    ymin=0,
    ymax=15,
    enlargelimits={abs=0.5},
    point meta=explicit,
    axis on top, ]
\nextgroupplot[
    yticklabel style = {rotate=-25,font=\tiny, anchor=east, inner sep=1pt},
    xticklabel style = {rotate=45,font=\tiny, anchor=north east, inner sep=1pt},
    colormap/viridis,
    colorbar,
    point meta min={0},
    point meta max={650},
    colorbar style={
            ytick={0, 200, 400, 650},
            width=0.05*\pgfkeysvalueof{/pgfplots/parent axis width},
        },
 	title style={at={(0.5,0)},anchor=north,yshift=-10mm},
 	title = FID,
    ]
\addplot [matrix plot*,mesh/cols=16,mesh/rows=16] file [meta=index 2] {data/exp/word2vec.langmap.fid.dat};		\nextgroupplot[
    yticklabel style = {rotate=-25,font=\tiny, anchor=east, inner sep=1pt},
    xticklabel style = {rotate=45,font=\tiny, anchor=north east, inner sep=1pt},
    colormap/viridis,
    colorbar,
    point meta min={0},
    point meta max={0.5},
    colorbar style={
            ytick={0, 0.25, 0.5},
            width=0.05*\pgfkeysvalueof{/pgfplots/parent axis width},
        },
 	title style={at={(0.5,0)},anchor=north,yshift=-10mm},
 	title = KID,
    ]
\addplot [matrix plot*,mesh/cols=16,mesh/rows=16] file [meta=index 2] {data/exp/word2vec.langmap.kid.dat};
  \end{groupplot}
\end{tikzpicture}
\caption{FID and KID are not able to capture language affinity from unaligned word2vec embeddings.}\label{fig:wordvec-map-fidkid}
\end{figure*}
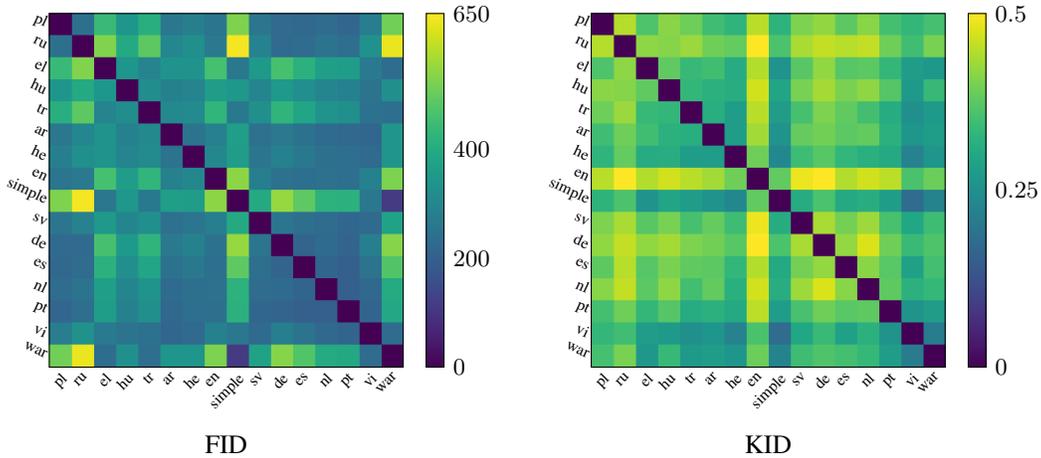

Figure \ref{fig:wordvec-map-fidkid} shows the matrix of distances for FID and KID aligned and colored in the same way as Figure \ref{fig:wordvec-multifig} (a).
FID and KID can not find meaningful structure in the data in the same way as \thiswork{} as they rely on extrinsic data properties.

\subsubsection{Word embedding experiment details.}
\begin{wrapfigure}[15]{r}{4.75cm}
\vspace{-15mm}
\centering
\begin{tikzpicture}
    \begin{axis}[xtick=\empty,ytick=\empty,ztick=\empty,width=6cm,height=4cm,view={-67.5}{67.5}]
        \addplot3[
            scatter/classes={
                a={mark=asterisk, cycle3, opacity=0.8},
                b={mark=asterisk, cycle2, opacity=0.6}
                },
                scatter, only marks,
                scatter src=explicit symbolic]
         table[x=x,y=y,z=z,meta=class]
            {data/torus.dat}; %
    \end{axis}
\end{tikzpicture}
\begin{center}
\small
\setlength{\tabcolsep}{1pt}
\renewcommand{\aboverulesep}{0pt}
\renewcommand{\belowrulesep}{0pt}
\newcolumntype{C}{>{\centering\arraybackslash}X}
\begin{tabularx}{5cm}{p{1cm}CC}
\multicolumn{1}{l}{\emph{metric}} & good GAN & bad GAN \\
\midrule
FID & $0.00529 \pm 0.00070$ & $0.00627 \pm 0.00076$ \\
KID & $0.00172 \pm 0.00073$ & $0.00259 \pm 0.00077$\\
\thiswork{} & $9.02059 \pm 1.5195$ & $\mathbf{14.0732} \pm 2.1706$ \\

\bottomrule
\end{tabularx}
\caption{
Bad GAN produces samples inside the torus hole \mbox{(red)}. FID and KID cannot detect such behaviour.
}
\label{fig:torus}
\end{center}
\end{wrapfigure} We use gensim~\citep{gensim} to learn word vectors on the latest Wikipedia corpus snapshot on 16 languages: Polish, Russian, Greek, Hungarian, Turkish, Arabic, Hebrew, English, Simple English, Swedish, German, Spanish, Dutch, Portugese, Vietnamese, and Waray-Waray. We then compute FID, KID and \thiswork{} scores on all the pairs, we average 100 runs for the heatmap figures~\ref{fig:wordvec-multifig}. For the different dimensionality experiment, we learn vectors on the English Wikipedia of sizes equal to the powers of 2 from 4 to 512. After that we compute \thiswork{} and covariance error, i.e. normalized PIP loss, between the pairs of sizes to generate the heatmap figure~\ref{fig:wordvec-map-diffdim-msid-pip}.

\subsubsection{Vanilla GAN on torus}
We provide an additional experiment clearly showing the case where \thiswork is superior to its main competitors, FID and KID. We train two vanilla GANs on the points of a 3D torus. The bad GAN fails to learn the topology of the dataset it tries to mimic, yet previous metrics cannot detect this fact. \thiswork{}, on the contrary, can tell the difference. Figure~\ref{fig:torus} shows the points sampled from the GAN with some of the points inside the hole. \mm{KID and FID confidence intervals overlap for good and bad GANs, meanwhile \thiswork{} scores are clearly distinct from each other.}

\subsubsection{Normalization Details}\label{sssec:normalization}
For the purpose of normalizing \thiswork{}, we need to approximate that graph's eigenvalues. \citet{coja2007} proved that $\lambda_1 \leq 1-c\bar{d}^{\nicefrac{-1}{2}} \leq \lambda_2 \leq \lambda_n \leq 1+c\bar{d}^{\nicefrac{-1}{2}}$ for the core of the graph for some constant $c$.
We have empirically found that $c=2$ provides a tight approximation for random graphs.
That coincides with the analysis of \citet{chung2004}, who proved that ${\lambda_n = (1+o(1))2\bar{d}^{\nicefrac{-1}{2}}}$ if ${d_{\mathrm{min}} \gg \sqrt{\bar{d}} \log^3 n}$ even though in our case $d_{\mathrm{min}} = \bar{d} = k$. We thus estimate the spectrum of a random Erd{\H{o}}s\hyp{}R{\'e}nyi graph as growing linearly between ${\lambda_1=1-2\bar{d}^{\nicefrac{-1}{2}}}$ and ${\lambda_n=1+2\bar{d}^{\nicefrac{-1}{2}}}$, which corresponds to the underlying manifold being two-dimensional~\citep{tsitsulin2018}.

\subsection{Experimental settings}\label{ssec:expset}

 We train all our models on a single server with NVIDIA V100 GPU with $16$Gb memory and $2~\times~20$ core Intel E5-2698~v4 CPU.
For the experiment summarized in Table~1 in the Section~4.1 we train WGAN and WGAN-GP models on 4 datasets: MNIST, FashionMNIST, CIFAR10 and CelebA and sample $10$k samples, $\mY$, from each of the GANs. We uniformly subsample $10$k images from the original datasets, $\mX$, and compute the \thiswork{}, KID and FID scores between $\mX$ and $\mY$. 
We report the mean as well as the 99\% confidence interval across 100 runs.

Below we report the architectures, hyperparameters and generated samples of the models used for the experiments.
We train each of the GANs for 200 epochs on MNIST, FMNIST and CIFAR-10, and for 50 epochs on CelebA dataset.
For WGAN we use RMSprop optimizer with learning rate of $5\times10^{-5}$.
For WGAN-GP we use Adam optimizer with learning rate of $10^{-4}$, $\beta_1=0.9, \beta_2=0.999$.

\subsection{Graph example}
\begin{wrapfigure}[10]{r}{6cm}
\vspace{-8mm}
\includegraphics[width=5cm]{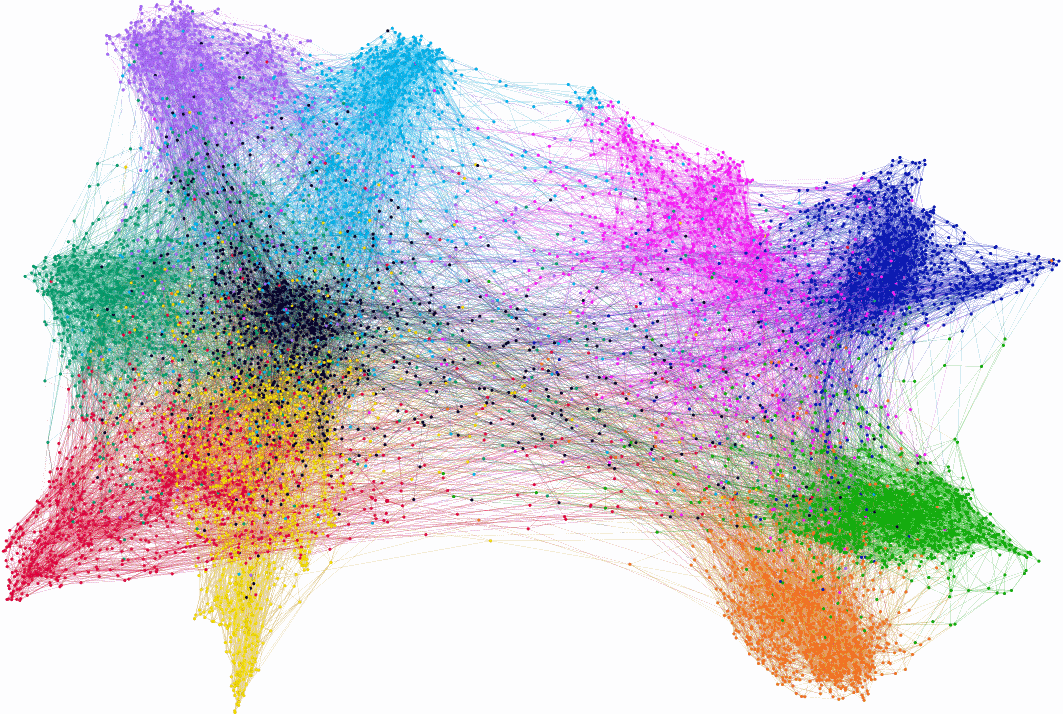}
\caption{CIFAR-10 graph colored with true class labels.}\label{fig:graph}
\end{wrapfigure}

Figure~\ref{fig:graph} provides visual proof that the $5\mathrm{NN}$ graph reflects the underlying manifold structure of the CIFAR-10 dataset. Clusters in the graph exactly correspond to CIFAR-10 classes.

\begin{figure}[p]
\includegraphics[width=0.5\textwidth]{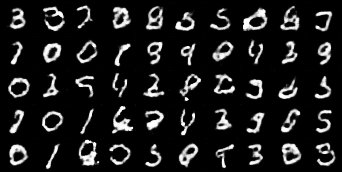}
\includegraphics[width=0.5\textwidth]{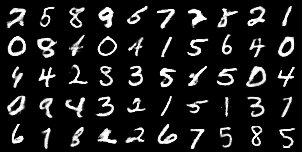}
\caption{MNIST samples (left: WGAN, right: WGAN-GP)}\label{fig:mnist}
\end{figure}

\begin{figure}[p]
\includegraphics[width=0.5\textwidth]{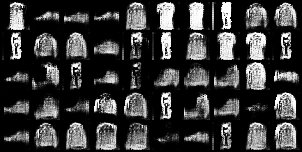}
\includegraphics[width=0.5\textwidth]{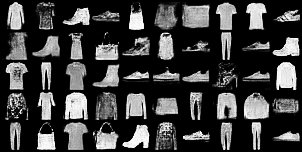}
\caption{FashionMNIST samples (left: WGAN, right: WGAN-GP)}\label{fig:fmnist}
\end{figure}

\begin{figure}[p]
\includegraphics[width=0.5\textwidth]{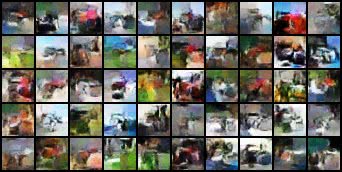}
\includegraphics[width=0.5\textwidth]{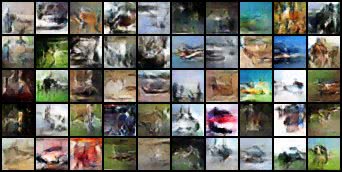}
\caption{CIFAR-10 samples (left: WGAN, right: WGAN-GP)}\label{fig:cifar10}
\end{figure}

\begin{figure}[p]
\includegraphics[width=0.5\textwidth]{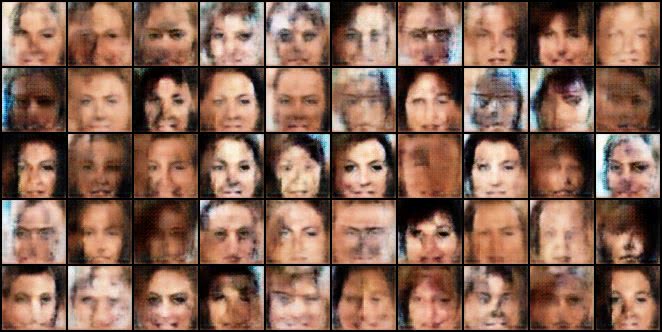}
\includegraphics[width=0.5\textwidth]{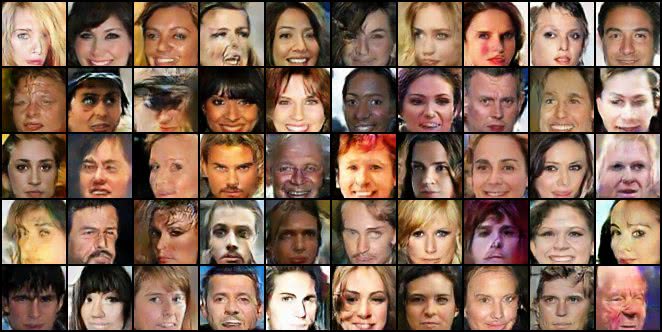}
\caption{CelebA samples (left: WGAN, right: WGAN-GP)}\label{fig:celeba}
\end{figure}

\begin{figure}[p]
\paragraph{MNIST WGAN}\ 
\begin{lstlisting}
ConvGenerator(
  (latent_to_features): Sequential(
    (0): Linear(in_features=100, out_features=512, bias=True)
    (1): ReLU()
  )
  (features_to_image): Sequential(
    (0): ConvTranspose2d(128, 64, kernel_size=(4, 4),
    			 stride=(2, 2), padding=(1, 1))
    (1): ReLU()
    (2): BatchNorm2d(64, eps=1e-05, momentum=0.1, affine=True)
    (3): ConvTranspose2d(64, 32, kernel_size=(4, 4),
    			 stride=(2, 2), padding=(1, 1))
    (4): ReLU()
    (5): BatchNorm2d(32, eps=1e-05, momentum=0.1, affine=True)
    (6): ConvTranspose2d(32, 16, kernel_size=(4, 4),
    			 stride=(2, 2), padding=(1, 1))
    (7): ReLU()
    (8): BatchNorm2d(16, eps=1e-05, momentum=0.1, affine=True)
    (9): ConvTranspose2d(16, 1, kernel_size=(4, 4), 
    			 stride=(2, 2), padding=(1, 1))
    (10): Sigmoid()
  )
)

ConvDiscriminator(
  (image_to_features): Sequential(
    (0): Conv2d(1, 16, kernel_size=(4, 4), stride=(2, 2), padding=(1, 1))
    (1): LeakyReLU(negative_slope=0.2)
    (2): Conv2d(16, 32, kernel_size=(4, 4), stride=(2, 2), padding=(1, 1))
    (3): LeakyReLU(negative_slope=0.2)
    (4): Conv2d(32, 64, kernel_size=(4, 4), stride=(2, 2), padding=(1, 1))
    (5): LeakyReLU(negative_slope=0.2)
    (6): Conv2d(64, 128, kernel_size=(4, 4), stride=(2, 2), padding=(1, 1))
    (7): Sigmoid()
  )
  (features_to_prob): Sequential(
    (0): Linear(in_features=512, out_features=1, bias=True)
    (1): Sigmoid()
  )
)
\end{lstlisting}
\end{figure}

\begin{figure}[p]
\paragraph{MNIST WGAN-GP, FMNIST (WGAN, WGAN-GP)}\ 
\begin{lstlisting}
MNISTGenerator(
  (block1): Sequential(
    (0): ConvTranspose2d(256, 128, kernel_size=(5, 5), stride=(1, 1))
    (1): ReLU(inplace)
  )
  (block2): Sequential(
    (0): ConvTranspose2d(128, 64, kernel_size=(5, 5), stride=(1, 1))
    (1): ReLU(inplace)
  )
  (deconv_out): ConvTranspose2d(64, 1, kernel_size=(8, 8), stride=(2, 2))
  (preprocess): Sequential(
    (0): Linear(in_features=128, out_features=4096, bias=True)
    (1): ReLU(inplace)
  )
  (sigmoid): Sigmoid()
)

MNISTDiscriminator(
  (main): Sequential(
    (0): Conv2d(1, 64, kernel_size=(5, 5), stride=(2, 2), padding=(2, 2))
    (1): ReLU(inplace)
    (2): Conv2d(64, 128, kernel_size=(5, 5), stride=(2, 2), padding=(2, 2))
    (3): ReLU(inplace)
    (4): Conv2d(128, 256, kernel_size=(5, 5), stride=(2, 2), padding=(2, 2))
    (5): ReLU(inplace)
  )
  (output): Linear(in_features=4096, out_features=1, bias=True)
)
\end{lstlisting}
\end{figure}

\begin{figure}[p]
\paragraph{CIFAR-10 (WGAN, WGAN-GP)}\ 
\begin{lstlisting}
CIFARGenerator(
  (preprocess): Sequential(
    (0): Linear(in_features=128, out_features=4096, bias=True)
    (1): BatchNorm1d(4096, eps=1e-05, momentum=0.1, affine=True)
    (2): ReLU(inplace)
  )
  (block1): Sequential(
    (0): ConvTranspose2d(256, 128, kernel_size=(2, 2), stride=(2, 2))
    (1): BatchNorm2d(128, eps=1e-05, momentum=0.1, affine=True)
    (2): ReLU(inplace)
  )
  (block2): Sequential(
    (0): ConvTranspose2d(128, 64, kernel_size=(2, 2), stride=(2, 2))
    (1): BatchNorm2d(64, eps=1e-05, momentum=0.1, affine=True)
    (2): ReLU(inplace)
  )
  (deconv_out): ConvTranspose2d(64, 3, kernel_size=(2, 2), stride=(2, 2))
  (tanh): Tanh()
)

CIFARDiscriminator(
  (main): Sequential(
    (0): Conv2d(3, 64, kernel_size=(3, 3), stride=(2, 2), padding=(1, 1))
    (1): LeakyReLU(negative_slope=0.01)
    (2): Conv2d(64, 128, kernel_size=(3, 3), stride=(2, 2), padding=(1, 1))
    (3): LeakyReLU(negative_slope=0.01)
    (4): Conv2d(128, 256, kernel_size=(3, 3), stride=(2, 2), padding=(1, 1))
    (5): LeakyReLU(negative_slope=0.01)
  )
  (linear): Linear(in_features=4096, out_features=1, bias=True)
)
\end{lstlisting}
\end{figure}

\begin{figure}[p]
\paragraph{CelebA (WGAN, WGAN-GP)}\ 
\begin{lstlisting}
CelebaGenerator(
  (preprocess): Sequential(
    (0): Linear(in_features=128, out_features=8192, bias=True)
    (1): BatchNorm1d(8192, eps=1e-05, momentum=0.1, affine=True)
    (2): ReLU(inplace)
  )
  (block1): Sequential(
    (0): ConvTranspose2d(512, 256, kernel_size=(5, 5), stride=(2, 2),
    			 padding=(2, 2), output_padding=(1, 1), bias=False)
    (1): BatchNorm2d(256, eps=1e-05, momentum=0.1, affine=True)
    (2): ReLU(inplace)
  )
  (block2): Sequential(
    (0): ConvTranspose2d(256, 128, kernel_size=(5, 5), stride=(2, 2),
    			 padding=(2, 2), output_padding=(1, 1), bias=False)
    (1): BatchNorm2d(128, eps=1e-05, momentum=0.1, affine=True)
    (2): ReLU(inplace)
  )
  (block3): Sequential(
    (0): ConvTranspose2d(128, 64, kernel_size=(5, 5), stride=(2, 2),
    			 padding=(2, 2), output_padding=(1, 1), bias=False)
    (1): BatchNorm2d(64, eps=1e-05, momentum=0.1, affine=True)
    (2): ReLU(inplace)
  )
  (deconv_out): ConvTranspose2d(64, 3, kernel_size=(5, 5), stride=(2, 2),
  				 padding=(2, 2), output_padding=(1, 1))
  (tanh): Tanh()
)
  
CelebaDiscriminator(
  (main): Sequential(
    (0): Conv2d(3, 64, kernel_size=(5, 5), stride=(2, 2), padding=(2, 2))
    (1): LeakyReLU(negative_slope=0.01)
    (2): Conv2d(64, 128, kernel_size=(5, 5), stride=(2, 2), padding=(2, 2))
    (3): LeakyReLU(negative_slope=0.01)
    (4): Conv2d(128, 256, kernel_size=(5, 5), stride=(2, 2), padding=(2, 2))
    (5): LeakyReLU(negative_slope=0.01)
    (6): Conv2d(256, 512, kernel_size=(5, 5), stride=(2, 2), padding=(2, 2))
    (7): LeakyReLU(negative_slope=0.01)
    (8): Conv2d(512, 1, kernel_size=(4, 4), stride=(1, 1))
  )
)
\end{lstlisting}

\end{figure}

 \end{document}